\documentclass[10pt,twocolumn,letterpaper]{article}

\usepackage{cvpr}              %

\usepackage{graphicx}
\usepackage{amsmath}
\usepackage{amssymb}
\usepackage{booktabs}

\usepackage[breaklinks,colorlinks]{hyperref}

\usepackage[capitalize]{cleveref}
\crefname{section}{Section}{Sections}
\Crefname{section}{Section}{Sections}
\Crefname{table}{Table}{Tables}
\crefname{table}{Table}{Tables}
\Crefname{figure}{Figure}{Figures}
\crefname{figure}{Figure}{Figures}

\def\cvprPaperID{21} %
\def\confName{CVPR}
\def\confYear{2023}

\usepackage{times}
\usepackage{epsfig}
\usepackage{graphicx}
\usepackage{booktabs}

\usepackage{csquotes}
\usepackage[english]{babel}
\usepackage[export]{adjustbox}
\usepackage{caption}
\usepackage{flushend}   %
\usepackage{subcaption}
\usepackage{lipsum}
\usepackage[all]{nowidow} %

\usepackage{placeins}%

\usepackage{array,multirow}

\usepackage{float}

\usepackage[acronym]{glossaries}

\usepackage{amsmath}
\usepackage{amsfonts}
\usepackage{amssymb}
\usepackage{amsthm}
\usepackage{enumerate}  %
\usepackage{subcaption}

\usepackage{hhline}

\usepackage{siunitx}
\newcommand{\chref}[2]{}

\usepackage[style=ieee,
    doi=false,
    url=false,
    isbn=false,
    mincitenames=1,
    maxcitenames=1,
    minbibnames=6,
    maxbibnames=6,
    date=year,  %
    backend=biber]{biblatex}  %

\DeclareSourcemap{
    \maps[datatype=bibtex, overwrite]{
        \map{
            \step[fieldset=address, null]
            \step[fieldset=location, null]
            \step[fieldset=note, null]
            \step[fieldset=series, null]
            \step[fieldset=issn, null]
            \step[fieldset=pages, null]
            \step[fieldset=editor, null]
            \step[fieldset=primaryclass, null]
            \step[fieldset=publisher, null]
            \step[fieldset=volume, null]
            \step[fieldset=issue, null]
            \step[fieldset=number, null]
            \step[fieldset=eventtitle, null]
        }
        \map[overwrite]{
            \pertype{online} %
            \step[fieldset=archiveprefix, null] %
            \step[fieldset=arxivid, null] %
            \step[fieldset=eprint, null] %
            \step[fieldset=pubstate, null] %
            \step[fieldset=urldate, null] %
        }
        \map{
            \step[fieldsource=booktitle, final] %
            \step[fieldset=archiveprefix, null] %
            \step[fieldset=arxivid, null] %
            \step[fieldset=eprint, null] %
        }
        \map{
            \step[fieldsource=journaltitle, final] %
            \step[fieldset=archiveprefix, null] %
            \step[fieldset=arxivid, null] %
            \step[fieldset=eprint, null] %
        }
    }
}

\AtEveryBibitem{\clearfield{month}}
\AtEveryBibitem{\clearfield{day}}

\renewcommand{\wrt}{w.r.t.\ } %

\newcommand{\idest}{i.e.\ } %
\renewcommand{\eg}{e.g.\ } %
\renewcommand{\cf}{cf.\ } %

\newcommand{\realworld}{real-world}
\newcommand{\meshrcnn}{Mesh {R-CNN}}
\newcommand{\maskrcnn}{Mask {R-CNN}}
\newcommand{\cubercnn}{Cube {R-CNN}}

\newcommand{\resnetfpn}{ResNet-50-FPN}
\newcommand{\dlafpn}{DLA-34-FPN}
\newcommand{\cuberefine}{{CubeRefine} {R-CNN}}

\newcommand{\dsparcel}{Parcel3D}

\newcommand{\ap}[1]{$\text{AP}_{#1}$}
\newcommand{\boxap}[1]{$\text{Box AP}_{#1}$}

\newcommand{\meshap}[1]{$\text{Mesh AP}_{#1}$}
\newcommand{\cubeap}[1]{$\text{AP3D}_{#1}$}
\newcommand{\cd}{Chamfer Distance}
\newcommand{\nc}{Normal Consistency}

\newcommand{\urlparcel}{\href{https://a-nau.github.io/parcel3d}{https://a-nau.github.io/parcel3d}}

\newglossaryentry{i4.0}{name={Industry 4.0}, description={Industry 4.0}}
\newglossaryentry{forklift}{name={forklift truck}, description={}, plural={forklift trucks}}
\newacronym[longplural={automated guided vehicles}]{agv}{AGV}{automated guided vehicle}
\newacronym{rfid}{RFID}{Radio-Frequency Identification}

\newacronym[longplural={time-of-flight cameras}]{tofc}{ToF camera}{time-of-flight camera}
\newacronym{pmd}{PMD}{Photonic Mixing Device}
\newacronym{roi}{ROI}{Region of Interest}
\newacronym{iou}{IoU}{Intersection over Union}
\newacronym{ar}{AR}{Augmented Reality}
\newacronym[longplural={Light Detection And Ranging}]{lidar}{LiDAR}{Light Detection And Ranging}
\newacronym[longplural={Frames per Second}]{fps}{FPS}{Frame per Second}
\newacronym{ocr}{OCR}{Optical Character Recognition}
\newacronym{gso}{GSO}{Google Scanned Objects}
\newacronym{lld}{LLD}{Large Logo Dataset}
\newacronym{rpn}{RPN}{Region Proposal Network}
\newacronym{fpn}{FPN}{Feature Pyramid Network}

\newacronym{AI}{AI}{Artificial Intelligence}
\newacronym{ML}{ML}{Machine Learning}
\newacronym{ransac}{RANSAC}{Random Sampling Consensus}
\newacronym[longplural={Artificial Neural Networks}]{nn}{ANN}{Artificial Neural Network}
\newacronym[longplural={Convolutional Neural Networks}]{cnn}{CNN}{Convolutional Neural Network}
\newacronym[longplural={Graph Convolutional Neural Networks}]{gcn}{GCN}{Graph Convolutional Neural Network}
\newacronym[longplural={Graph Neural Networks}]{gnn}{GNN}{Graph Neural Network}
\newacronym[longplural={Support Vector Machines}]{svm}{SVM}{Support Vector Machine}
\newacronym{dbscan}{DBSCAN}{Density-Based Spatial Clustering of Applications with Noise}
\newacronym{sgdm}{SGD+M}{Stochastic Gradient Descent with Momentum}
\newacronym{hmm}{HMM}{Hidden Markov Model}
\newacronym{hci}{HCI}{Human-Computer-Interaction}

\newacronym{eu}{EU}{European Union}

\newglossaryentry{poc}{name={proof of concept}, description={}}
\newglossaryentry{sota}{name={state-of-the-art}, description={}}

\newcommand{\thresholdChamf}{d_{cham}}
\newcommand{\thresholdNormal}{c_{norm}}
\newcommand{\dPicked}{$\mathbb{M}_{\text{Pick}}$}
\newcommand{\dRemoved}{$\mathbb{M}_{\text{Rem}}$}
\newcommand{\dDistractor}{$\mathbb{M}_{\text{Distr}}$} 

\makeglossaries
\begin{document}

\title{Parcel3D: Shape Reconstruction from Single RGB Images for Applications in Transportation Logistics
}

\makeatletter
\let\@fnsymbol\@arabic
\makeatother

\author{
    Alexander Naumann,  %
    Felix Hertlein, %
    Laura D\"orr %
    and 
    Kai Furmans \\
    ~\\
    FZI Research Center for Information Technology, Karlsruhe, Germany
    and \\
    Karlsruhe Institute of Technology (KIT),
    Karlsruhe, Germany\\%
    ~\\
    \tt\{anaumann, hertlein, doerr\}@fzi.de ~ kai.furmans@kit.edu
}
\maketitle

\begin{abstract}

    We focus on enabling damage and tampering detection in logistics and tackle the problem of 3D shape reconstruction of potentially damaged parcels.
    As input we utilize single RGB images, which corresponds to use-cases where only simple handheld devices are available, \eg for postmen during delivery or clients on delivery.
    We present a novel synthetic dataset, named \dsparcel{}, that is based on the \gls*{gso} dataset and consists of more than 13,000 images of parcels with full 3D annotations.
    The dataset contains intact, \idest cuboid-shaped, parcels and damaged parcels, which were generated in simulations.
    We work towards detecting mishandling of parcels by presenting a novel architecture called \cuberefine{},
    which combines estimating a 3D bounding box with an iterative mesh refinement.
    We benchmark our approach on \dsparcel{} and an existing dataset of cuboid-shaped parcels in real-world scenarios.
    Our results show, that while training on \dsparcel{} enables transfer to the real world, enabling reliable deployment in real-world scenarios is still challenging.
    \cuberefine{} yields competitive performance in terms of \meshap{} and is the only model that directly enables deformation assessment by 3D mesh comparison and tampering detection by comparing viewpoint invariant parcel side surface representations.
    Dataset and code are available at \urlparcel{}.
\end{abstract}

\section{Introduction}
\label{introduction}

Transportation logistics and warehousing are a central part of every supply chain and play an important strategic role in the Industry 4.0 era \cite{tangStrategicRoleLogistics2019}.
However, several challenges need to be faced by companies working in the logistics sector: clients demand cheaper, faster and more precisely scheduled deliveries while at the same time, cities and highways are congested, and environmental concerns are of rising importance.
To tackle these challenges, process automation has huge potential \cite{woschankReviewFurtherDirections2020}.
Key processes for automation in logistics are identification, digital measurement, damage detection and tampering recognition of packaging units, all of which we work towards with the approach presented in this work.
Identification is necessary for process documentation and parcel tracking. %
Damage and tampering detection can be utilized to increase the safety and security along the supply chain \cite{nocetiMulticameraSystemDamage2018}.
Finally, digital measurement and volume estimation are essential for the optimization of vessel capacity usage \cite{zhaoComparativeReview3D2016}.

\begin{figure}[t!]
    \centering
    \includegraphics[width=\linewidth]{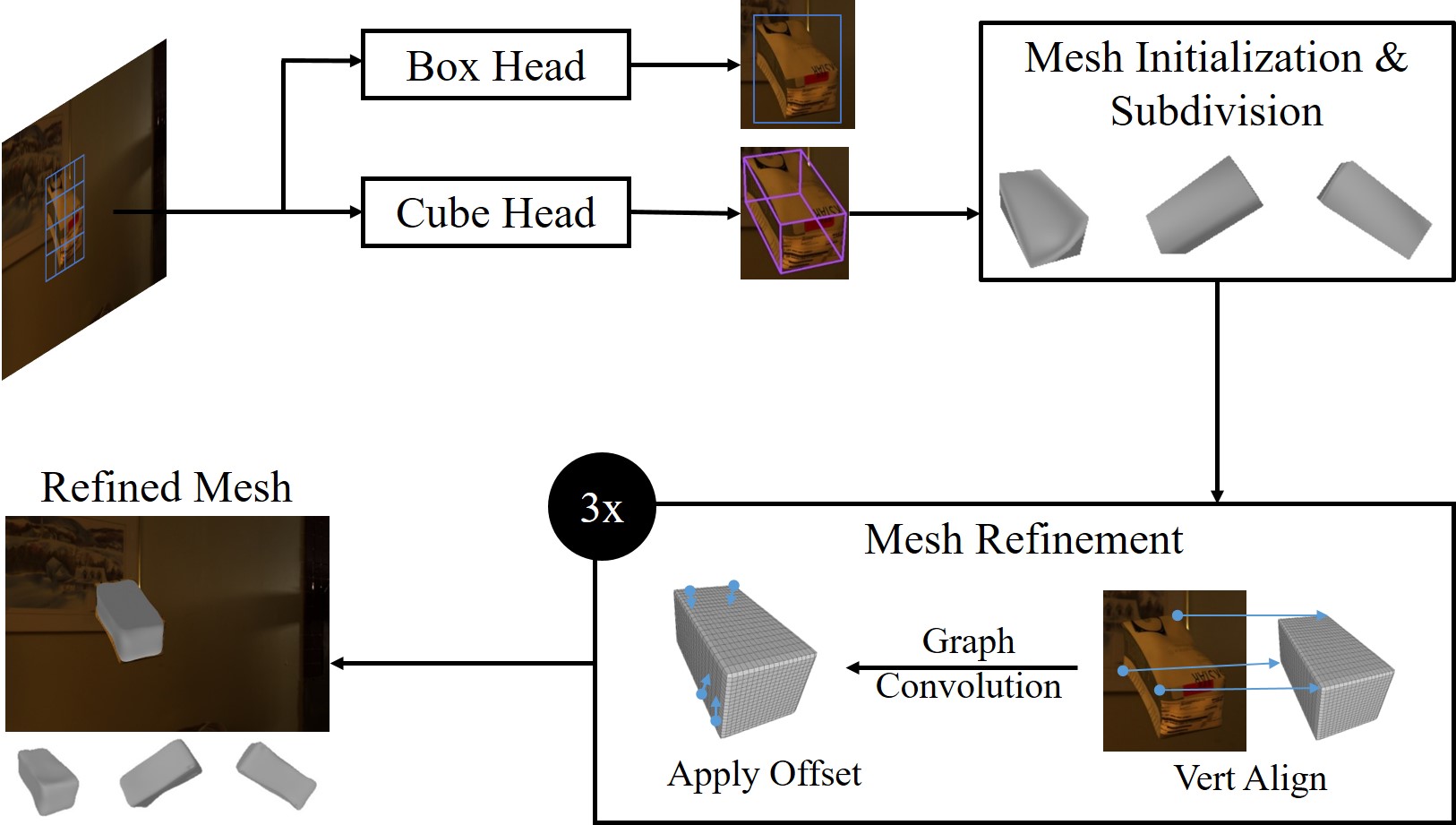}
    \caption{
        We take an RGB image as input and use \cubercnn{}'s \textit{Cube Head} \cite{brazilOmni3DLargeBenchmark2023} to estimate a 3D bounding box.
        This bounding box is subdivided and serves as initial mesh, which is refined by an iterative mesh refinement as proposed in \cite{wangPixel2MeshGenerating3D2018}.
        For training and evaluation we present \dsparcel{}, a novel dataset of normal and damaged parcels with full 3D annotations.
    }
    \label{fig:overview}
    
\end{figure}

Before introducing our key ideas, we shortly present important features of the logistics domain, since these characteristics influence our dataset and architecture design decisions.
In logistics, packaging is usually used to handle, transport and store goods in a safe and efficient way \cite{saghirConceptPackagingLogistics2004}.
The most common choice are cuboid-shaped packaging units.
Moreover, transportation logistics is a constrained environment, with a lot of standardizations:
Loading devices, such as the EPAL Euro pallet\footnote{See \href{https://cn.epal-pallets.org/fileadmin/user_upload/ntg_package/images/mediathek/DU_GB_EPAL_1_Produktdatenblatt_low.pdf}{https://www.epal-pallets.org}.}, or standardized labels, such as GS1 STILL\footnote{See \href{https://www.gs1.org/standards/gs1-logistic-label-guideline/1-3}{https://www.gs1si.org}.}, to name a few examples.
Finally, due to the ubiquity of logistics processes in manufacturing businesses and beyond, it is crucial to develop flexible and easy to use solutions.
In fact, a study by \textcite{nocetiMulticameraSystemDamage2018} showed, that easy integration of novel automated processes such as damage detection, is a crucial factor for technology adoption.

In this work, we present an approach for the automation of localization and shape reconstruction in logistics, which is outlined in \cref{fig:overview}.
We focus on the detection and shape reconstruction for intact (cuboid-shaped) and damaged parcels.
By leveraging the standardizations mentioned before, \idest by using objects with known sizes as references, length measurements are also possible with monocular images. %
We prioritize flexibility and thus, refrain from using expensive sensors or even multi-sensor setups.
Instead, our approach solely relies on a single RGB image as input.
Since RGB cameras are already integrated into many handheld devices that are used in transportation logistics, our approach is suitable for various scenarios where high flexibility is needed, such as for postmen during delivery or for clients on delivery.

There has been active research in the area of single image 3D reconstruction \cite{choy3DR2N2UnifiedApproach2016a, wangPixel2MeshGenerating3D2018,}.
We use \cubercnn{} \cite{brazilOmni3DLargeBenchmark2023} as base architecture, since a 3D bounding box closely describes cuboid-shaped parcels and yields a suitable shape-prior for damaged parcels.
We extend \cubercnn{} by an iterative mesh refinement as used in \cite{wangPixel2MeshGenerating3D2018,gkioxariMeshRCNN2019}.
This enables us to (1) leverage a strong prior as starting point for the mesh refinement process and (2) it simultaneously estimates the original shape of parcels in form of the 3D bounding box and their potentially deformed current state.
The latter enables a direct comparison between 3D meshes for damage quantification and tampering detection by comparing viewpoint invariant parcel side surface representations \cite{nocetiMulticameraSystemDamage2018}.
One important issue with 3D reconstruction, however, is the availability of suitable datasets, which are scarce due to the excessive annotation costs.
Thus, most approaches are trained on synthetic data or data for specific domains (\eg Pix3D \cite{sunPix3DDatasetMethods2018}), %
however, no suitable dataset in the area of transportation logistics exists.
To overcome this, we introduce the new synthetic dataset \dsparcel{}, that is built by automatically selecting suitable \glsfirst*{gso} \cite{downsGoogleScannedObjects2022} models and generating damaged parcel models through simulation with Blender\footnote{See \href{https://www.blender.org/}{https://www.blender.org/}.}.
We employ a flexible rendering pipeline that includes varying camera parameters, lighting and scene contexts.
Since the parcel texture is crucial for rendering realistic images, we also present a small synthetic cardboard texture dataset.
In addition, we use a real dataset of intact parcels for evaluation and 
report a \boxap{} of up to $82.1$ and \meshap{50} of $32.3$, which confirms the suitability of our synthetic dataset for applications in logistics.
Note, that due to the lack of suitable datasets we do not evaluate damage pattern recognition and tampering detection.

The main contributions of this work are:
\begin{itemize}
    \item we introduce \dsparcel{}, a novel synthetic dataset of intact and damaged parcel images with full 3D annotations that allows transfer to real images, and
    \item we present \cuberefine{}, a novel architecture targeting single image 3D reconstruction for applications in transportation logistics, which combines 3D bounding box estimation with an iterative mesh refinement.
\end{itemize}
This work is structured as follows.
In \cref{sec:related_work} we present an overview of related literature.
\cref{sec:data} outlines the dataset generation and \cref{sec:architecture} our novel neural network architecture.
\cref{sec:results_and_evaluation} evaluates our approach on synthetic and real data, and \cref{sec:conclusion} concludes the paper.
\section{Related Work}
\label{sec:related_work}

To the best of our knowledge there is no prior work on shape reconstruction from single images in transportation logistics and warehousing.
We review literature on applications in logistics, cuboid reconstruction from RGB images and finally, 3D reconstruction of arbitrary objects from single images in the following.

\paragraph*{Applications in Logistics.}
There is work on 2D segmentation of parcels \cite{naumannRefinedPlaneSegmentation2020,naumannScrapeCutPasteLearn2022}, packaging units \cite{mayershoferFullySyntheticTrainingIndustrial2020,mayershoferLOCOLogisticsObjects2020} and packaging structure recognition \cite{dorrFullyAutomatedPackagingStructure2020,dorrTetraPackNetFourCornerBasedObject2021}.
Moreover, there has been research on 3D reconstruction from RGBD images \cite{liUsingKinectMonitoring2012,prasseNewApproachesSingularization2015,sonMethodConstructAutomatic2017,arpentiRGBDRecognitionLocalization2020} and from multiple views \cite{nocetiMulticameraSystemDamage2018}.
3D reconstruction by using RFID technology has been explored in \cite{bu3DimensionalReconstructionTagged2017}.
Damage and tampering detection has been tackled by \textcite{nocetiMulticameraSystemDamage2018} in a constrained multi-camera setup.
Tampering is detected by comparing normalized parcel side surfaces and damage detection by fitting a parallelepiped across multiple views.
For an in-depth review on computer vision applications in logistics, we refer to \textcite{naumannLiteratureReviewComputer2023}.

\paragraph*{Cuboid reconstruction.}
Cuboid reconstruction from single RGB images by identifying its 8 corner points in 2D has been tackled in the literature.
Approaches are class agnostic, meaning that diverse object categories are considered as either cuboid or not.
\textcite{xiaoLocalizing3DCuboids2012} present such an approach in the pre-deep learning era that leverages corner and edge detection techniques.
After the rise of deep learning, also cuboid reconstruction was tackled with \glspl*{nn}.
\textcite{dwibediDeepCuboidDetection2016} present an approach to estimate the position of the 8 cuboid keypoints using deep learning.
A similar line of work is concerned with 3D bounding box estimation for cars \cite{fang3DBoundingBox2019,liuGroundawareMonocular3D2021,kumarDEVIANTDepthEquiVarIAnt2022}, which is reviewed in-depth by \textcite{ma3DObjectDetection2022}.
Note, that by assuming that cars are driving on the road, rotation estimation can be reduced to yaw estimation.
Approaches leverage geometric priors by requiring consistent vanishing points \cite{ruiGeometryConstrainedCarRecognition2020} and by imposing 2D/3D consistency \cite{liRTM3DRealTimeMonocular2020}.
Recently, \textcite{brazilOmni3DLargeBenchmark2023} introduced a large benchmark for 3D object detection, which combines several existing datasets.
Moreover, they present a simple and effective model for 3D object detection, called \cubercnn{}.

\paragraph*{Single RGB image 3D reconstruction.}
There are many approaches for general image-based 3D reconstruction without a confinement to an object type.
While the input for many approaches is a single RGB image, the output varies: representations based on voxels \cite{choy3DR2N2UnifiedApproach2016a, xiePix2VoxContextaware3D2019,yangSingleView3DObject2021},
meshes \cite{kanazawaLearningCategorySpecificMesh2018,wenPixel2MeshMultiView3D2019,gkioxariMeshRCNN2019,}
and pointclouds \cite{fanPointSetGeneration2017, gadelhaMultiresolutionTreeNetworks2018} are common.
In addition to that, implicit representations \cite{meschederOccupancyNetworksLearning2019,zakharovSingleShotSceneReconstruction2021}  have been introduced.
Most reconstruction approaches focus on single instances, either by considering only images with a single instance or by employing 2D segmentation.
More recently, also NeRFs \cite{mildenhallNeRFRepresentingScenes2020} have been used to tackle single-view reconstruction \cite{mullerAutoRFLearning3D2022}.
Apart from supervised approaches, there has been work on 3D reconstruction from 2D supervision \cite{kanazawaLearningCategorySpecificMesh2018}, unpaired image collections \cite{duggalTopologicallyAwareDeformationFields2022} and unsupervised reconstruction \cite{insafutdinovUnsupervisedLearningShape2018,navaneetImageCollectionsPoint2020,wuUnsupervisedLearningProbably2020,huSelfSupervised3DMesh2021}, since training data with ground truth 3D annotations is difficult and costly to obtain.
\textcite{hanImagebased3DObject2021} present an overview of approaches from the deep learning era that leverage either single or multiple RGB images for 3D reconstruction.
The reviews of \textcite{fuSingleImage3D2021} and \textcite{khanThreeDimensionalReconstructionSingle2022} focus explicitly on single image 3D reconstruction.

~\\
We introduce the new dataset \dsparcel{} to enable research on image-based 3D reconstruction in the domain of logistics.
Furthermore, we leverage the existing general 3D object detection architecture \cubercnn{} \cite{brazilOmni3DLargeBenchmark2023} and extend it by an iterative mesh refinement.
Adding the iterative mesh refinement is necessary, since 3D object detection approaches are not suitable for damage detection and analysis.
In contrast to other 3D reconstruction approaches, \cuberefine{} directly enables comparing the original shape of a cuboid-shaped object with its current state, which is crucial for damage quantification.

\section{Synthetic Dataset: \dsparcel{}}
\label{sec:data}

We present details on the generation of our synthetic dataset \dsparcel{} and
start by describing the automatic selection process for suitable \gls*{gso} \cite{downsGoogleScannedObjects2022} object models in \cref{sec:data:models}.
Next, the approaches to generate data for damaged parcels and for new textures are presented in \cref{sec:data:blender} and \cref{sec:data:texture}, respectively. 
Finally, we present details on the rendering in \cref{sec:data:details}.

\subsection{Model Selection}
\label{sec:data:models}
\begin{figure}[h]
    \centering
    \begin{tabular}{ccc}
        Picked (\dPicked) & Removed (\dRemoved{}) & Distractor (\dDistractor{}) \\ 
        \hline
        \\
        \adjincludegraphics[width=0.13\linewidth, trim={{0.1\width} 0 {0.1\width} 0}, clip]{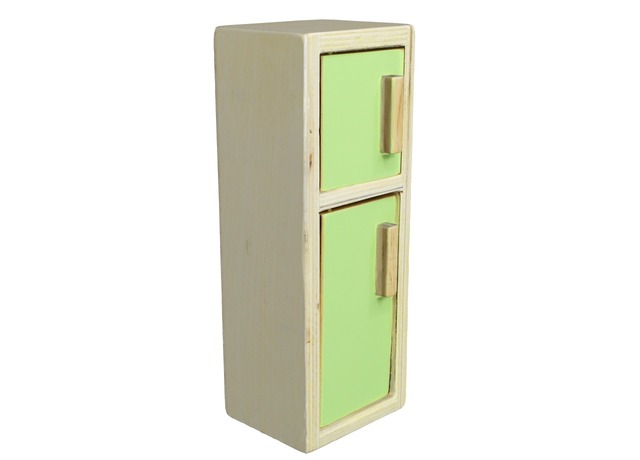}%
        \adjincludegraphics[width=0.13\linewidth, trim={{0.1\width} 0 {0.1\width} 0}, clip]{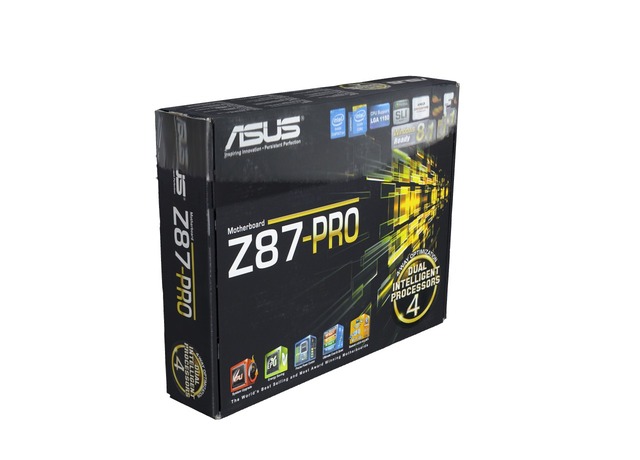}%
                          & 
        \adjincludegraphics[width=0.13\linewidth, trim={{0.1\width} 0 {0.1\width} 0}, clip]{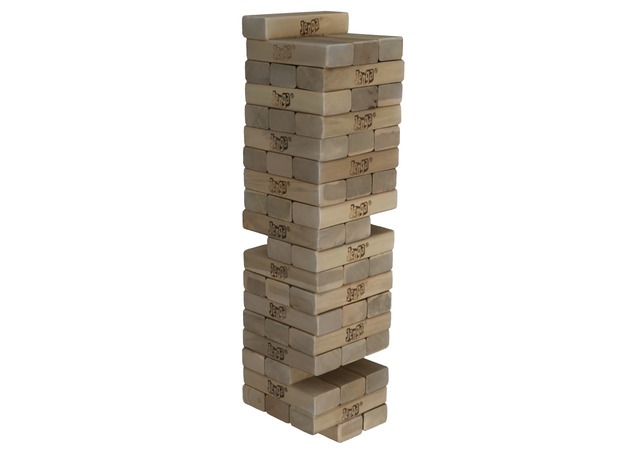}%
        \adjincludegraphics[width=0.13\linewidth, trim={{0.1\width} 0 {0.1\width} 0}, clip]{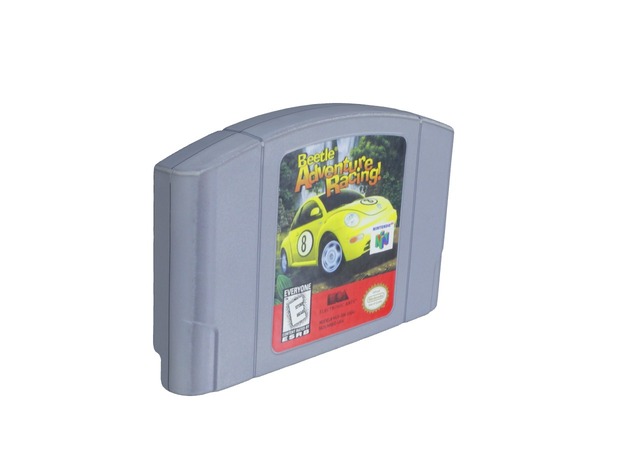}%
                          & 
        \adjincludegraphics[width=0.13\linewidth, trim={{0.1\width} 0 {0.1\width} 0}, clip]{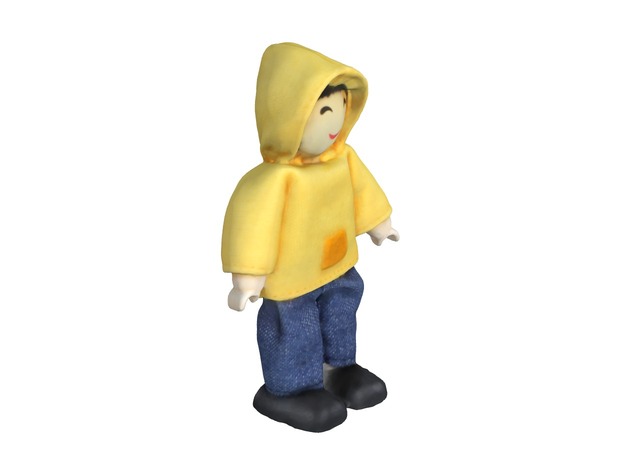}%
        \adjincludegraphics[width=0.13\linewidth, trim={{0.1\width} 0 {0.1\width} 0}, clip]{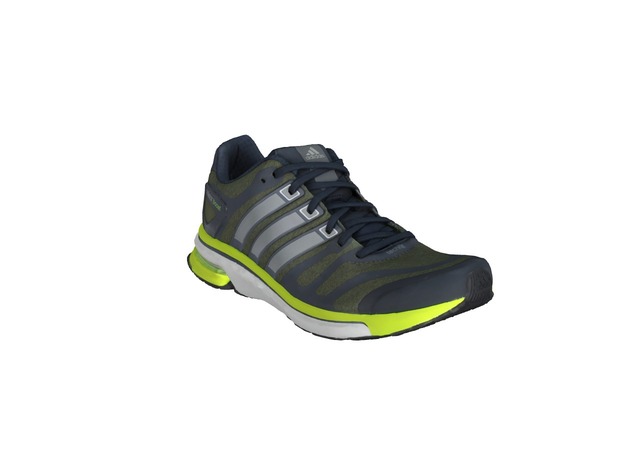}%
        \\
        \adjincludegraphics[width=0.13\linewidth, trim={{0.1\width} 0 {0.1\width} 0}, clip]{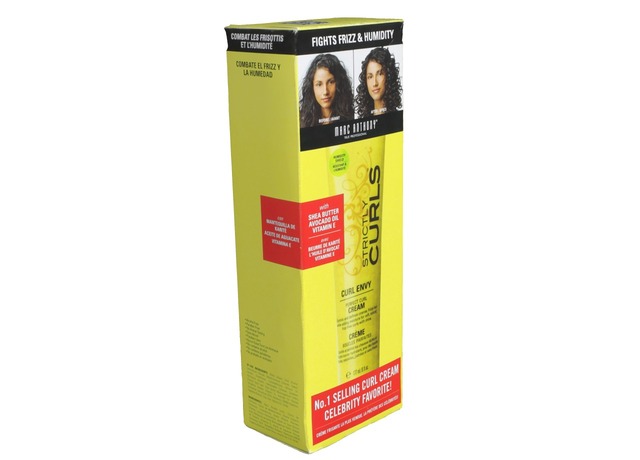}%
        \adjincludegraphics[width=0.13\linewidth, trim={{0.1\width} 0 {0.1\width} 0}, clip]{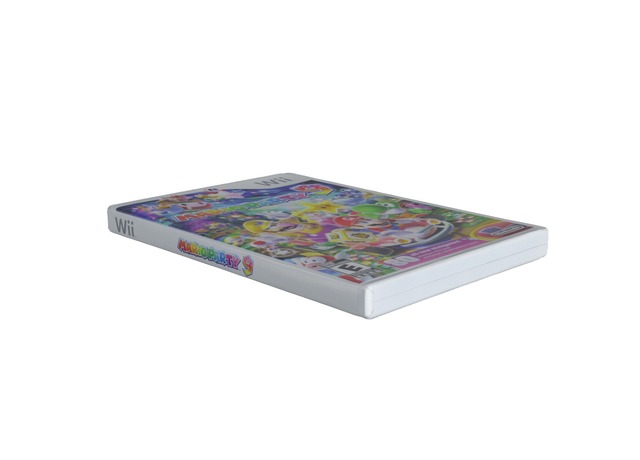}%
                          & 
        \adjincludegraphics[width=0.13\linewidth, trim={{0.1\width} 0 {0.1\width} 0}, clip]{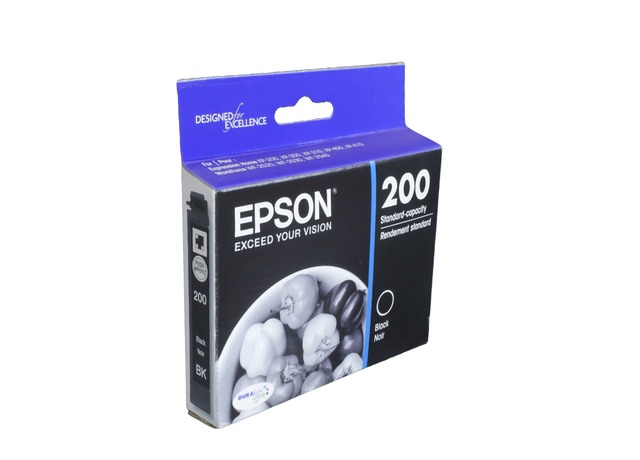}%
        \adjincludegraphics[width=0.13\linewidth, trim={{0.1\width} 0 {0.1\width} 0}, clip]{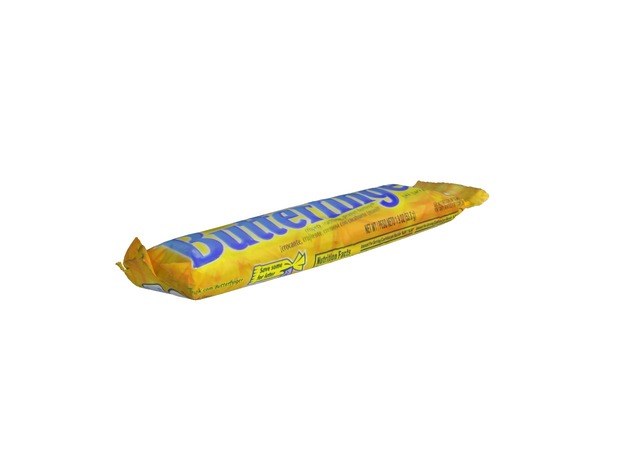}%
                          & 
        \adjincludegraphics[width=0.13\linewidth, trim={{0.1\width} 0 {0.1\width} 0}, clip]{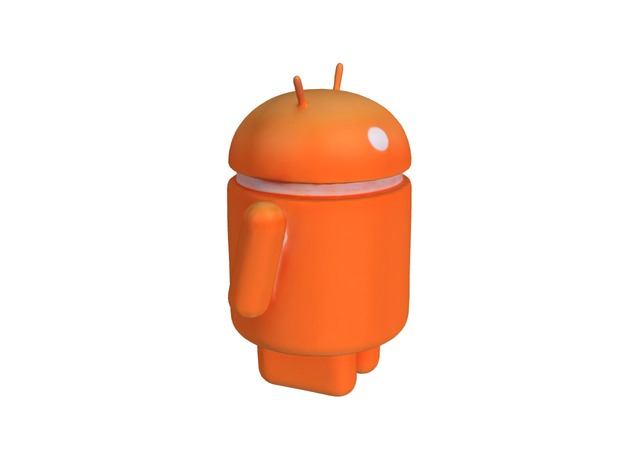}%
        \adjincludegraphics[width=0.13\linewidth, trim={{0.1\width} 0 {0.1\width} 0}, clip]{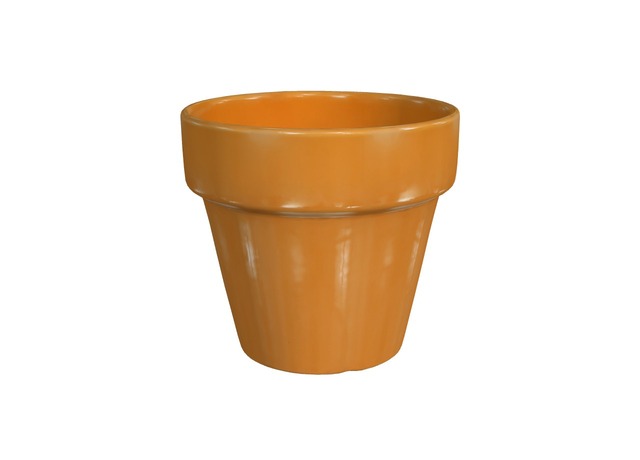}%
        \\
        \adjincludegraphics[width=0.13\linewidth, trim={{0.1\width} 0 {0.1\width} 0}, clip]{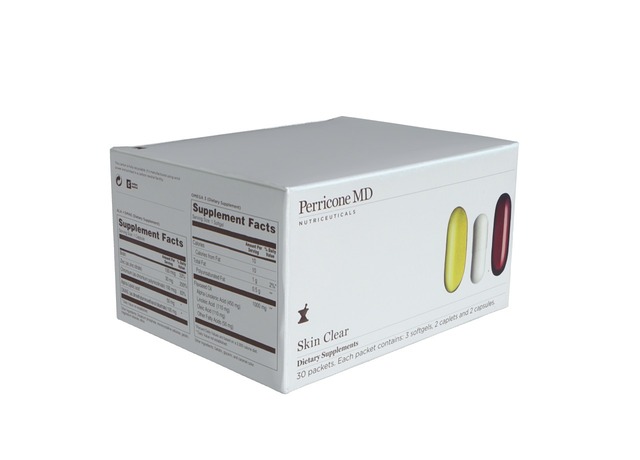}%
        \adjincludegraphics[width=0.13\linewidth, trim={{0.1\width} 0 {0.1\width} 0}, clip]{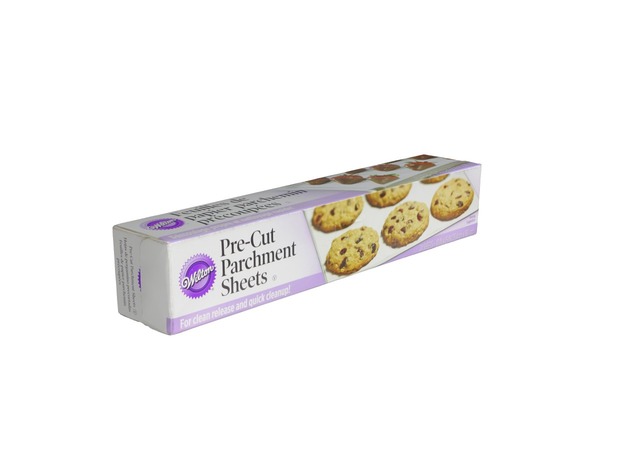}%
                          & 
        \adjincludegraphics[width=0.13\linewidth, trim={{0.1\width} 0 {0.1\width} 0}, clip]{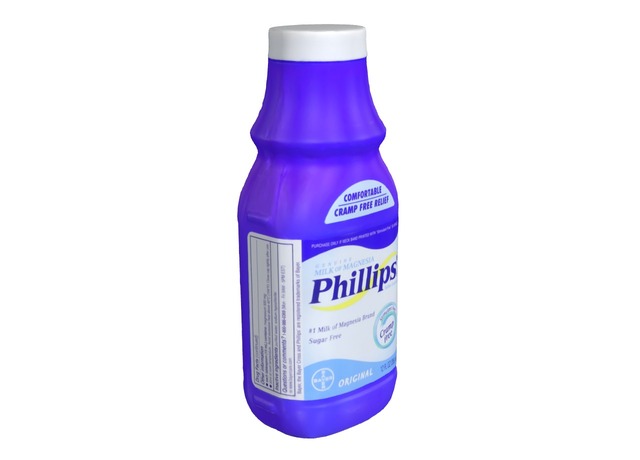}%
        \adjincludegraphics[width=0.13\linewidth, trim={{0.1\width} 0 {0.1\width} 0}, clip]{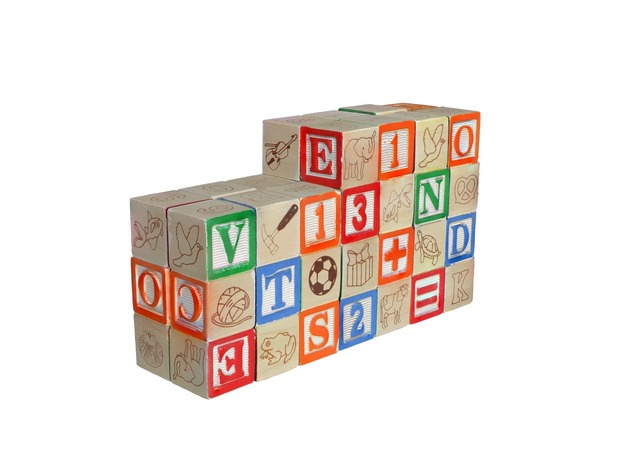}%
                          & 
        \adjincludegraphics[width=0.13\linewidth, trim={{0.1\width} 0 {0.1\width} 0}, clip]{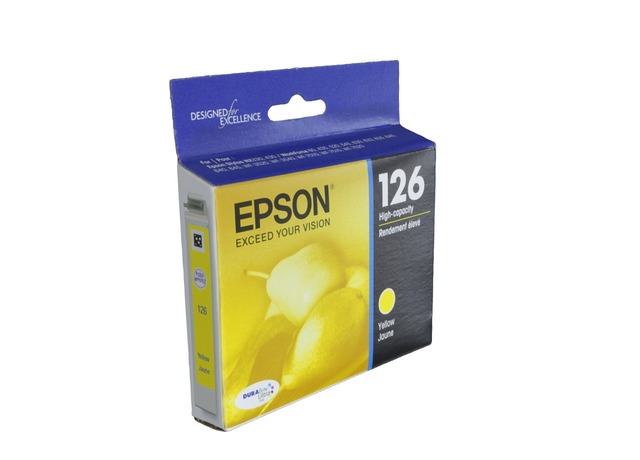}%
        \adjincludegraphics[width=0.13\linewidth, trim={{0.1\width} 0 {0.1\width} 0}, clip]{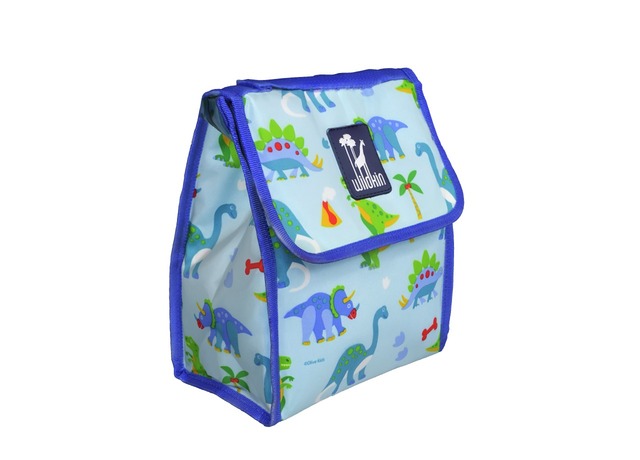}%
        \\
    \end{tabular}
    \caption{
        Samples of the three object model subsets of the \gls*{gso} dataset \cite{downsGoogleScannedObjects2022} that were generated based on the models' similarity with a cuboid.
    }
    \label{fig:model_selection}
\end{figure}

We use \gls*{gso} as a base dataset, since it has a wide variety of realistic 3D models.
We create a new subset of the \gls*{gso} dataset that is tailored towards our use-case in transportation logistics and warehousing by automatically selecting relevant models based on their shape.
This filtering is done by evaluating each model's similarity with a surrounding cuboid.
We initialize a template mesh from the surrounding cuboid and use the \cd{} $\thresholdChamf$ and \nc{} $\thresholdNormal$ between this template mesh and the model mesh for comparison.
~\\
We divide the models in three categories using empirically determined thresholds for both similarity metrics. 
Models with $\thresholdChamf \le 0.1$ and $\thresholdNormal \ge 0.9$ are chosen as cuboid models due to their high resemblance with the desired shape.
We refer to these picked models by \dPicked{}.
The second threshold of $\thresholdChamf \le 0.5$ and $\thresholdNormal \ge 0.8$ identifies objects that are not closely related to a cuboid in shape, yet similar.
These models are denoted \dRemoved{}.
All other models are referred to by \dDistractor{}.
We use models from \dDistractor{} as distractor objects, which we also render into images to prevent overfitting on rendering artifacts \cite{dwibediCutPasteLearn2017}.
The models from \dRemoved{} are not used as distractors, since their resemblance in shape with a cuboid might be confusing.
The subset \dDistractor{} contains 750 models, \dRemoved{} contains 71 models and \dPicked{} contains 209 models.
Exemplary instances for each of the three categories are visualized in \cref{fig:model_selection}.

\begin{figure}[h]
    \centering
    \adjincludegraphics[width=0.23\linewidth, trim={{0.1\width} 0 {0.1\width} 0}, clip]{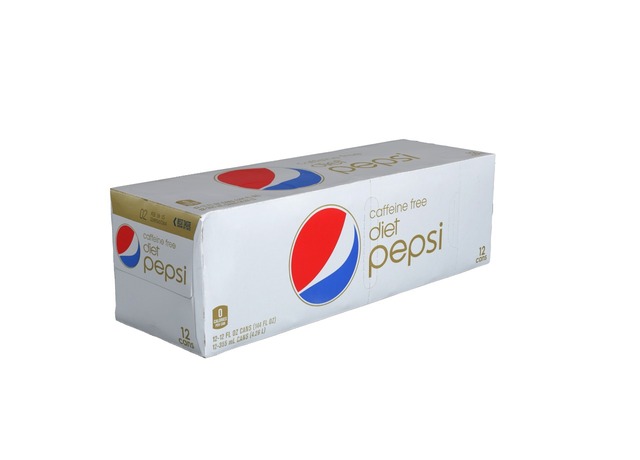}~
    \adjincludegraphics[width=0.23\linewidth, trim={{0.1\width} 0 {0.1\width} 0}, clip]{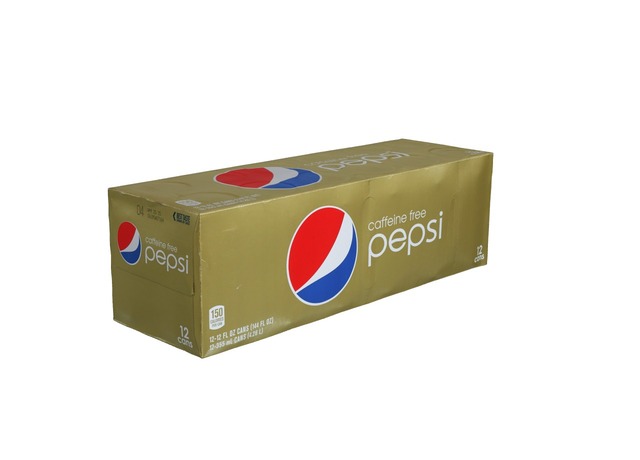}~
    \adjincludegraphics[width=0.23\linewidth, trim={{0.1\width} 0 {0.1\width} 0}, clip]{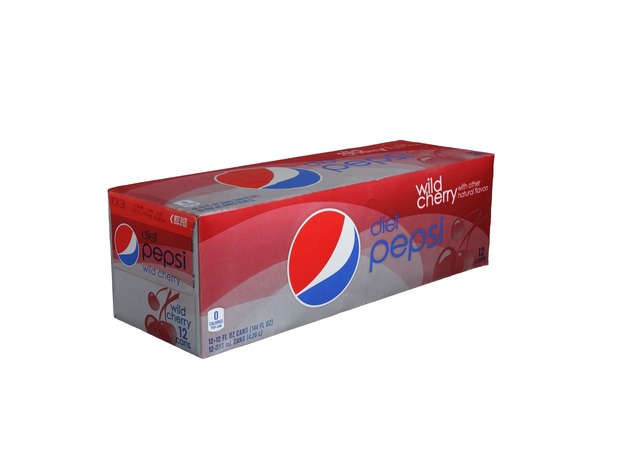}~
    \adjincludegraphics[width=0.23\linewidth, trim={{0.1\width} 0 {0.1\width} 0}, clip]{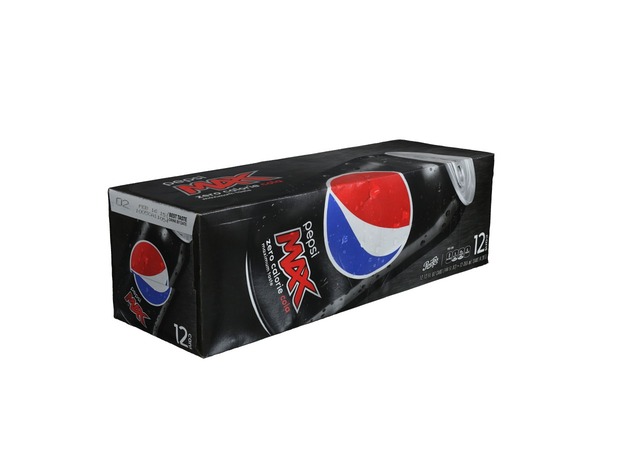}~
    \caption{
        Visualization of the similarity between certain models.
    }
    \label{fig:model_selection:grouping}
\end{figure}

~\\
Since there are very similar models within \dPicked{}, we combine the models into 66 groups.
The grouping is done automatically by using brand and category names, since the \gls*{gso} dataset contains similar object models as seen in \cref{fig:model_selection:grouping} for the example of Pepsi cartons.

\subsection{Model Generation}
\label{sec:data:blender}

\begin{figure}[h]
    \centering
    \adjincludegraphics[width=0.23\linewidth, trim={{0.25\width} {0.1\height} {0.25\width} {0.4\height}}, clip]{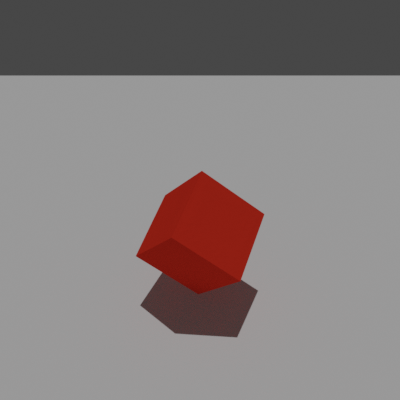}~
    \adjincludegraphics[width=0.23\linewidth, trim={{0.25\width} {0.1\height} {0.25\width} {0.4\height}}, clip]{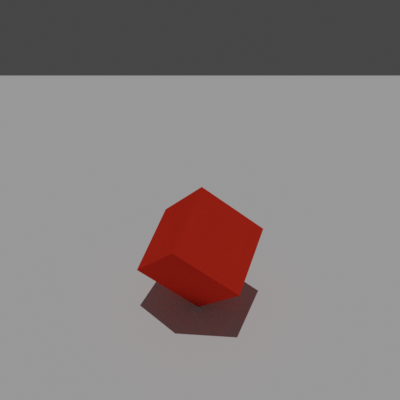}~
    \adjincludegraphics[width=0.23\linewidth, trim={{0.25\width} {0.1\height} {0.25\width} {0.4\height}}, clip]{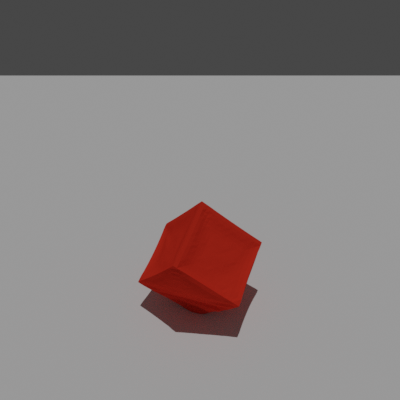}~
    \adjincludegraphics[width=0.23\linewidth, trim={{0.25\width} {0.1\height} {0.25\width} {0.4\height}}, clip]{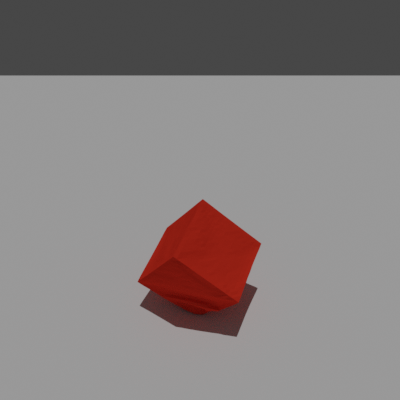}~
    \caption{Visualization of the collision for damaged parcels using soft body simulation in Blender.}
    \label{fig:blender}
\end{figure}

Since we obtain only 209 suitable models from the \gls*{gso} dataset, we generate 10 scaled versions for each of them. 
The scaling is done for each of the three dimensions separately by sampling a scaling factor from a triangular distribution with lower limit $0.5$, upper limit $2$ and mode $1$.
These models make up the subset of intact boxes.
~\\
This method for dataset generation is suitable for intact parcel recognition, however, automatically identifying suitable models for damaged boxes within the \gls*{gso} or other datasets such as ShapeNet \cite{changShapeNetInformationRich3D2015}, is difficult.
Thus, we automatically generate models for damaged boxes using physics-based simulation in Blender.
For each simulation, we start by randomly sampling a base model from the previously generated subset of intact boxes.
The chosen model is then simulated to be falling onto a rigid ground as seen in \cref{fig:blender}.
Soft body simulation is used to allow deformations during the collision.
We sample falling height, angle and soft body physics parameters randomly within empirically determined ranges to obtain a wide variety of deformations.
Only models from timesteps that have between $75\%$ and $90\%$ %
of their original volume are chosen as suitable models for damaged parcels.
These thresholds ensure that models have at least a certain degree of deformation, while not allowing extreme changes in appearance.
Furthermore, we use a RANSAC algorithm \cite{fischlerRandomSampleConsensus1987} to find the best rigid transformation between the original, cuboid-shaped model and the deformed model during simulation, to track the position and rotation of the object. 
Note, that this is necessary, since Blender does not incorporate the tracking of objects during a soft body simulation.
Using this information we are able to identify the area of impact with the strongest deformation, which allows us to render damaged parcels such that the impacted area is visible.
Finally, we apply a smoothing filter in Blender to the selected models.

\subsection{Texture Generation}
\label{sec:data:texture}

\begin{figure}[h]
    \begin{subfigure}[b]{0.33\linewidth}
        \centering
        \includegraphics[width=0.97\linewidth]{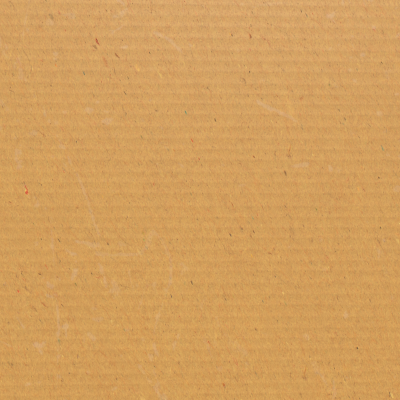}
        \caption{Cardboard}
        \label{fig:textures:a}
    \end{subfigure}%
    \begin{subfigure}[b]{0.33\linewidth}
        \centering
        \includegraphics[width=0.92\linewidth]{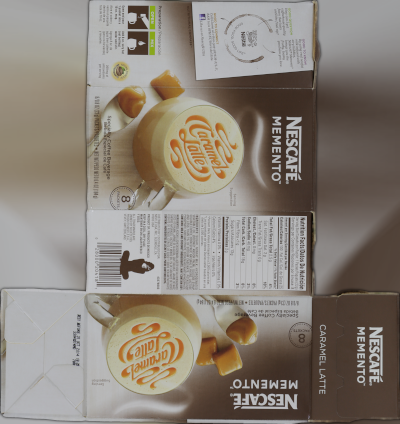}
        \caption{Extrapolation}
        \label{fig:textures:b}
    \end{subfigure}%
    \begin{subfigure}[b]{0.33\linewidth}
        \centering
        \includegraphics[width=0.97\linewidth]{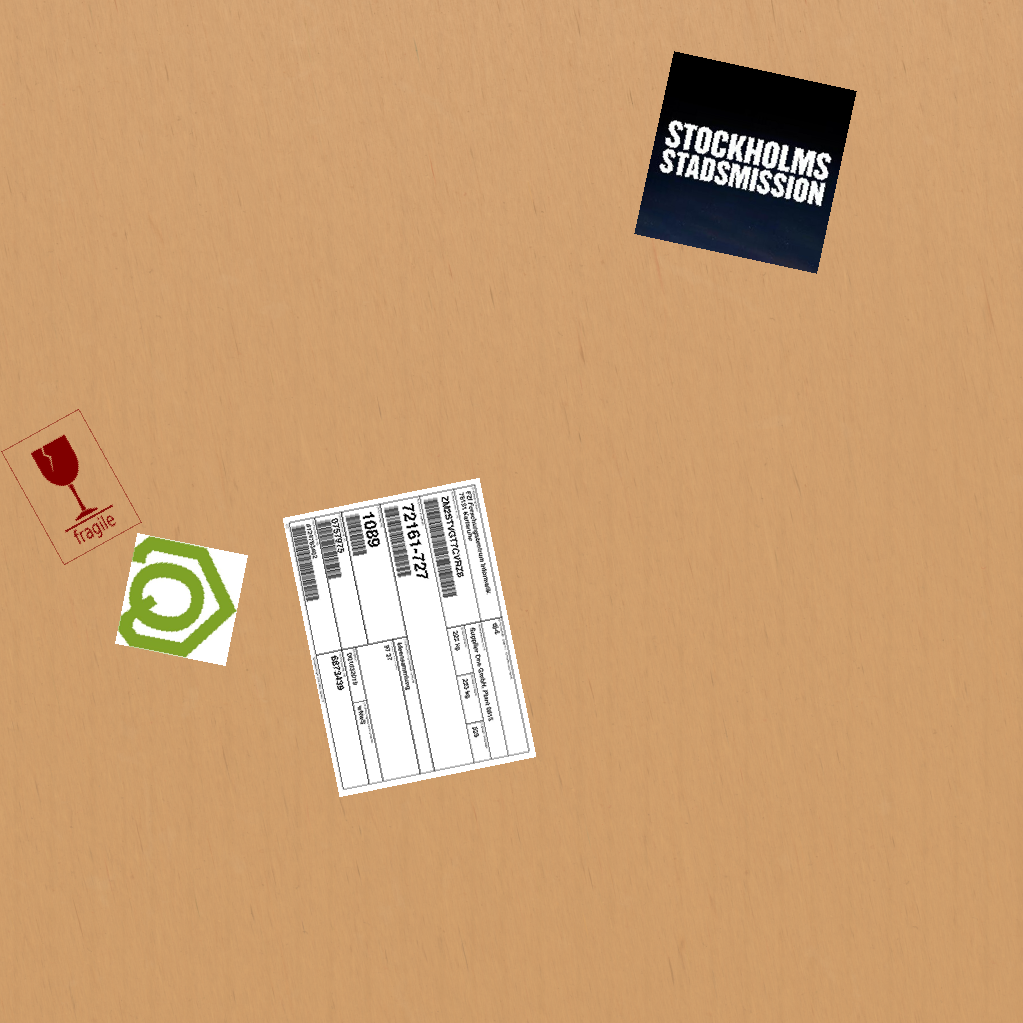}
        \caption{Labels \& Logos}
        \label{fig:textures:c}
    \end{subfigure}
    \caption{
        Examples for generated textures:
        (a) Plain cardboard texture, (b) extrapolation of existing textures for damaged parcels and (c) cardboard texture with labels and logos.
    }
    \label{fig:textures}
\end{figure}

In order to obtain more variance in the textures of the models and to bias the training data towards cardboard, we generate new textures.
We use a cardboard shader in Blender\footnote{See \href{https://blendermarket.com/products/cardboard}{https://blendermarket.com/products/cardboard}.}, to generate a dataset of 230 cardboard textures.
These textures replace the original texture of the model with a probability of $0.6$, and an example is shown in \cref{fig:textures:a}.
When the original texture is used for a damaged parcel, texture mapping is not trivial and we need to extrapolate the texture image.
This extrapolation is done using pixel-wise nearest neighbor averaging and an exemplary result can be seen in \cref{fig:textures:b}.
In addition, we randomly add to each texture 
\begin{itemize}
    \item 0-3 logos from the \gls*{lld} \cite{sage2017logodataset}
    \item 1 shipping label from a mix of 30 labels from \cite{dorrLeanTrainingData2019} and 65 labels found online, with a probability of $0.6$
    \item 0-2 fragile labels from 16 labels found online, with a probability of $0.4$
\end{itemize}
An example for a final cardboard texture with labels and logos is visualized in \cref{fig:textures:c}.

\subsection{Rendering Details}
\label{sec:data:details}

We sample 200 models randomly for each of the 66 groups, yielding more than \SI{13000}{} scenes, which we render with $1080 \times 720$ resolution.
Damaged models and cuboid-shaped models, respectively, are sampled with a probability of $50\%$ and textures are generated as described before.
We add $0$-$3$ randomly sampled distractor models from \dDistractor{} to the scene and use environment maps from \citeauthor{gardnerLearningPredictIndoor2017} \cite{gardnerLearningPredictIndoor2017} for realistic scene contexts.
We permit an occlusion of up to $30\%$ of the model of interest and generate a new image composition if the criteria is not met.

All assets that were used follow a $0.7$, $0.15$, $0.15$ split between training, validation and test data.
These splits were respected in the generation of the rendered images.
To have realistic poses of the objects we restrict the elevation angle to lie between $20^{\circ}$ and $60^{\circ}$ degrees.
The azimuth angle is sampled freely for intact and between $-30^{\circ}$ and $30^{\circ}$ degrees for deformed models, such that the damage is visible and not self-occluded.
We add small random rotations to the \textit{lookat} configuration resulting from azimuth and elevation angle and vary the focal length slightly at random.

\section{Approach}
\label{sec:architecture}

We present our novel model architecture \cuberefine{} that is targeted towards reconstructing potentially deformed cuboid objects such as parcels in \cref{sec:architecture:base}.
Furthermore, we present details on our training procedure in \cref{sec:architecture:training}.

\subsection{Neural Network Architecture}
\label{sec:architecture:base}

Our model \cuberefine{} extends \cubercnn{} \cite{brazilOmni3DLargeBenchmark2023} by adding an iterative mesh refinement (\cf \cref{fig:overview}).
\cubercnn{} is a general architecture that combines 2D detection with 3D bounding box estimation.
Its architecture consist of a backbone network for feature extraction, which is followed by a \gls*{rpn} \cite{renFasterRCNNRealTime2017}.
We follow the original work and use a DLA-34-FPN \cite{yuDeepLayerAggregation2018,linFeaturePyramidNetworks2017} as backbone.
The generated region proposals are then passed on to two different branches.
The first branch is a \textit{Box Head}, which outputs a 2D bounding box and the category label.
The second branch estimates the 3D bounding box and is called \textit{Cube Head}.
It takes $7\times7$ feature maps pooled from the region-aligned backbone features and passes them to two fully connected layers with hidden dimension 1024.
A final fully connected layer predicts 13 parameters which represent the 3D bounding box.
Note, that this architecture could be easily extended to encompass a full \maskrcnn{} \cite{heMaskRCNN2017} by adding segmentation.
For details, we refer to \textcite{brazilOmni3DLargeBenchmark2023}.

For the mesh refinement, we extend the \textit{Cube Head} by subdividing its 8-point mesh triangulation output four times to obtain an initial mesh prediction of sufficient granularity.
Note, that without the iterative subdivision, the mesh representation would be too coarse to accurately represent parcel deformations.
The subdivided mesh is then passed on to the mesh refinement stage.
We follow \textcite{gkioxariMeshRCNN2019}, and use three refinement stages with three graph convolutions each.
In each stage, image features from the backbone are aligned with the vertices of the current mesh version and graph convolutions are applied to compute a positional offset for each vertex in the mesh.
These mesh offsets should morph the current mesh representation such that the mesh closely depicts the real parcel shape.
We experimented with different options for message passing within the graph such as Residual Gated Graph Convolution \cite{bressonResidualGatedGraph2018}, EG \cite{tailorWeNeedAnisotropic2022} and GATv2 \cite{brodyHowAttentiveAre2022}.
Since no significant improvements were observed, we stick to the original architecture.

\cuberefine{} leverages a cuboid prior, which is a valid assumption for both cuboid-shaped and most damaged parcels.
Compared to \meshrcnn{}, the \textit{Cube Head} is more lightweight than the \textit{Voxel Head}.
Moreover, our model predicts both, the original shape of the parcel and the possibly deformed current shape of the parcel at the same time.
We discuss the advantages of this in more detail in \cref{sec:eval:summary}.

\subsection{Training Procedure}
\label{sec:architecture:training}

We follow the same training procedure for all our training runs.
We choose a batch size of 16, use \gls*{sgdm} with a base learning rate of $0.02$.
The learning rate increases linearly from $0.002$ over the first \SI{1500}{} iterations.
Subsequently, we divide the learning rate by four in iterations \SI{7500}{}, \SI{12500}{} and \SI{17500}{}.
The maximum number of iterations is set to \SI{20000}{}.

During our experiments, we consider two different backbones, namely a ResNet-50  \cite{heDeepResidualLearning2016} and a DLA-34 \cite{yuDeepLayerAggregation2018}, both in combination with a \gls*{fpn} \cite{linFeaturePyramidNetworks2017}.
We freeze the backbone weights at stage four and initialize them using pre-trained weights from \textcite{gkioxariMeshRCNN2019} and \textcite{brazilOmni3DLargeBenchmark2023}.

\section{Evaluation}
\label{sec:results_and_evaluation}

In the following, we present our evaluation of 2D bounding box detection, 3D bounding box detection and shape reconstruction on synthetic and real data.
Due to the lack of annotated real data of damaged parcels, the quantitative \realworld{} evaluation only presents results on cuboid-shaped parcels.
We benchmark our model against Pix2Mesh \cite{wangPixel2MeshGenerating3D2018}\footnote{We use the implementation of \textcite{gkioxariMeshRCNN2019}.}, \meshrcnn{} \cite{gkioxariMeshRCNN2019} and \cubercnn{} \cite{brazilOmni3DLargeBenchmark2023} by training and evaluating on the respective splits of \dsparcel{}.
Unless stated otherwise, we use the same \dlafpn{} backbone and three mesh refinement stages with three graph convolutions each, to enable a direct comparison between approaches.
We present results for the original version of \meshrcnn{} with a \resnetfpn{} backbone, however, focus on the comparable results in the following.

All results are summarized in \cref{table:eval:mesh} and \cref{table:eval:cube}, and we present details on the evaluation for synthetic data in \cref{sec:eval:synthetic} and for real data in \cref{sec:eval:real}.
Finally, we summarize the findings focusing on the \realworld{} applicability in \cref{sec:eval:summary}.

\begin{table*}[t]
    
    \centering
    \small
    \begin{tabular}{lllccccccc}
        \toprule
                                                                     &  &                        & \multicolumn{3}{c}{Box} & \multicolumn{2}{c}{Mesh} & Chamfer             & Normal                                                                                       \\ 
        Model                                                        &  & Dataset                & \ap{}                   & \ap{50}                  & \ap{75}             & \ap{50}             & \ap{75}             & Distance ($\downarrow$) & Consistency            \\
        \midrule
        Pix2Mesh \cite{wangPixel2MeshGenerating3D2018}               &  & Intact                 & 96.0 (0.5)              & 98.9 (0.5)               & 98.7 (0.2)          & 89.6 (1.0)          & 48.5 (1.6)          & 0.311 (0.086)           & 0.901 (0.001)          \\
        Mesh R-CNN (RN50) \cite{gkioxariMeshRCNN2019} \hspace{-1cm}  &  & Intact                 & 93.2 (0.4)              & 98.3 (0.3)               & 97.9 (0.3)          & 82.9 (1.2)          & 42.6 (1.1)          & 1.924 (1.214)           & 0.886 (0.001)          \\
        Mesh R-CNN \cite{gkioxariMeshRCNN2019}                       &  & Intact                 & 95.9 (0.5)              & 98.9 (0.5)               & 98.7 (0.2)          & \textbf{92.9 (1.6)} & 67.0 (2.8)          & 0.225 (0.088)           & 0.914 (0.001)          \\
        Cube R-CNN \cite{brazilOmni3DLargeBenchmark2023}             &  & Intact                 & 97.1 (0.1)              & \textbf{99.0 (0.0)}      & \textbf{99.0 (0.0)} & 92.0 (0.3)          & 74.4 (2.0)          & 0.159 (0.016)           & 0.925 (0.001)          \\
        \cuberefine{} (ours)\hspace{-0.75cm}                         &  & Intact                 & \textbf{97.1 (0.0)}     & \textbf{99.0 (0.0)}      & \textbf{99.0 (0.0)} & {92.8 (0.2)}        & \textbf{77.2 (1.2)} & \textbf{0.128 (0.002)}  & \textbf{0.929 (0.001)} \\
        \midrule
        Pix2Mesh \cite{wangPixel2MeshGenerating3D2018}               &  & Damaged\hspace{-0.2cm} & 95.1 (0.6)              & \textbf{99.8 (0.1)}      & 98.8 (0.2)          & 84.3 (1.2)          & 12.4 (1.4)          & 0.750 (0.553)           & 0.866 (0.002)          \\
        Mesh R-CNN (RN50) \cite{gkioxariMeshRCNN2019}  \hspace{-1cm} &  & Damaged\hspace{-0.2cm} & 92.1 (0.4)              & 99.6 (0.1)               & 98.9 (0.4)          & 78.8 (0.7)          & 9.0 (0.4)           & 0.599 (0.322)           & 0.859 (0.001)          \\
        Mesh R-CNN \cite{gkioxariMeshRCNN2019}                       &  & Damaged\hspace{-0.2cm} & 94.6 (0.5)              & 99.2 (0.5)               & 98.8 (0.3)          & \textbf{91.1 (0.5)} & \textbf{26.1 (1.9)} & 0.860 (0.436)           & \textbf{0.880 (0.002)} \\
        Cube R-CNN \cite{brazilOmni3DLargeBenchmark2023}             &  & Damaged\hspace{-0.2cm} & 95.0 (0.2)              & 99.0 (0.0)               & \textbf{99.0 (0.0)} & 32.6 (0.5)          & 0.1 (0.0)           & 0.494 (0.004)           & 0.806 (0.000)          \\
        \cuberefine{} (ours)\hspace{-0.75cm}                         &  & Damaged\hspace{-0.2cm} & \textbf{95.2 (0.1)}     & 99.0 (0.0)               & \textbf{99.0 (0.0)} & 70.7 (0.7)          & 4.1 (0.2)           & \textbf{0.293 (0.003)}  & 0.861 (0.000)          \\
        \midrule
        Pix2Mesh \cite{wangPixel2MeshGenerating3D2018}               &  & Real                   & 74.4 (1.9)              & 93.4 (1.7)               & 89.3 (2.4)          & 27.8 (2.1)          & 2.3 (0.6)           & 2.112 (0.060)           & 0.744 (0.006)          \\
        Mesh R-CNN (RN50) \cite{gkioxariMeshRCNN2019} \hspace{-1cm}  &  & Real                   & \textbf{82.1 (0.7)}     & \textbf{99.0 (0.0)}      & \textbf{97.8 (0.1)} & 32.0 (0.4)          & 5.0 (1.0)           & 1.965 (0.050)           & 0.756 (0.002)          \\
        Mesh R-CNN \cite{gkioxariMeshRCNN2019}                       &  & Real                   & 70.6 (5.0)              & 89.2 (5.9)               & 84.4 (5.7)          & 29.4 (2.7)          & 4.9 (1.5)           & 2.153 (0.073)           & 0.742 (0.008)          \\
        Cube R-CNN \cite{brazilOmni3DLargeBenchmark2023}             &  & Real                   & 43.4 (6.9)              & 52.8 (8.3)               & 49.9 (7.3)          & 30.1 (5.8)          & \textbf{13.3 (4.3)} & 0.875 (0.041)           & 0.808 (0.003)          \\
        \cuberefine{} (ours)\hspace{-0.75cm}                         &  & Real                   & 41.5 (5.8)              & 50.3 (6.6)               & 47.6 (6.5)          & \textbf{32.3 (4.2)} & 13.1 (3.0)          & \textbf{0.814 (0.062)}  & \textbf{0.828 (0.006)} \\
        \bottomrule
    \end{tabular}
    \vspace{0.1cm}
    \caption{
        Quantitative performance analysis of mesh reconstruction on different datasets.
        The \meshap{} is the mean area under the Precision-Recall curve for {F1@0.3$>x$}, as in \cite{gkioxariMeshRCNN2019}.
        We repeated all trainings five times and report mean values with standard deviations in parentheses.
        The best mean performance for each dataset type is highlighted.
    }
    \label{table:eval:mesh}
    
\end{table*}

\begin{table*}[t]
    
    \centering
    \begin{tabular}{llccc}
        \toprule
        Model                                            & Dataset & \cubeap{}  & \cubeap{15} & \cubeap{25} \\
        \midrule
        Cube R-CNN \cite{brazilOmni3DLargeBenchmark2023} & Intact  & 69.5 (0.8) & 81.6 (1.1)  & 74.4 (1.4)  \\
        \cuberefine{} (ours)                             & Intact  & 69.3 (0.6) & 80.9 (0.5)  & 74.1 (1.1)  \\
        \midrule
        Cube R-CNN \cite{brazilOmni3DLargeBenchmark2023} & Damaged & 86.6 (0.3) & 94.4 (0.6)  & 89.9 (0.8)  \\
        \cuberefine{} (ours)                             & Damaged & 86.5 (0.6) & 94.6 (0.6)  & 89.7 (0.6)  \\
        \midrule
        Cube R-CNN \cite{brazilOmni3DLargeBenchmark2023} & Real    & 53.3 (8.6) & 53.8 (8.7)  & 53.8 (8.7)  \\
        \cuberefine{} (ours)                             & Real    & 50.6 (6.8) & 51.1 (6.8)  & 51.1 (6.8)  \\
        \bottomrule
    \end{tabular}
    \vspace{0.1cm}
    \caption{
        Quantitative performance analysis of 3D object detection for Cube R-CNN and \cuberefine{} on different datasets.
        The average precision for 3D IoU (\cubeap{}) is computed as in \cite{brazilOmni3DLargeBenchmark2023}.
        We repeated all trainings five times and report mean values with standard deviations in parentheses.
    }
    \label{table:eval:cube}
    
\end{table*}

\subsection{Synthetic Data}
\label{sec:eval:synthetic}

We consider the case of intact parcels and damaged parcels separately by evaluating only on the respective subsets of the \dsparcel{} test dataset.
The performance for 2D bounding box detection is very high for all models on our presented synthetic dataset \dsparcel{} with the lowest observed \boxap{} being $92.1$ (\cf \cref{table:eval:mesh}).

\begin{figure}
    \centering
    \begin{subfigure}[b]{.47\linewidth}
        \centering
        \includegraphics[width=\linewidth]{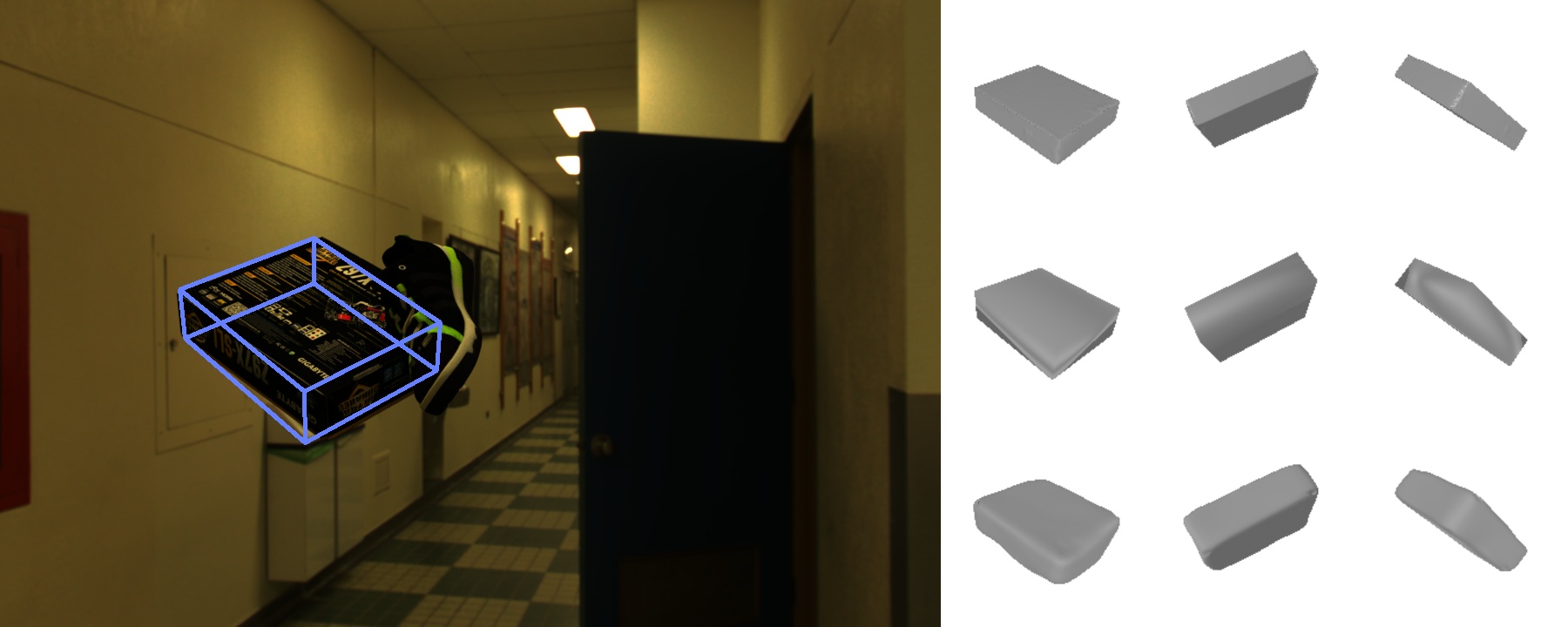}
        \includegraphics[width=\linewidth]{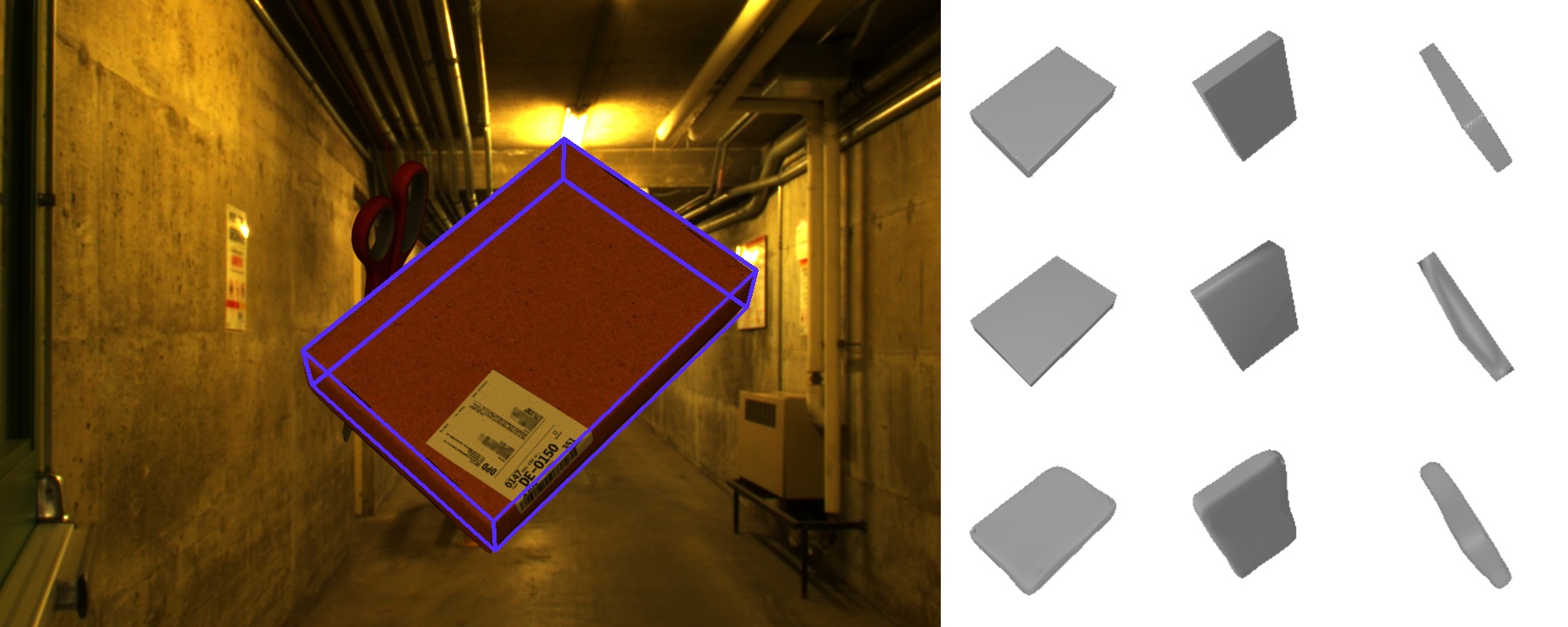}
        \includegraphics[width=\linewidth]{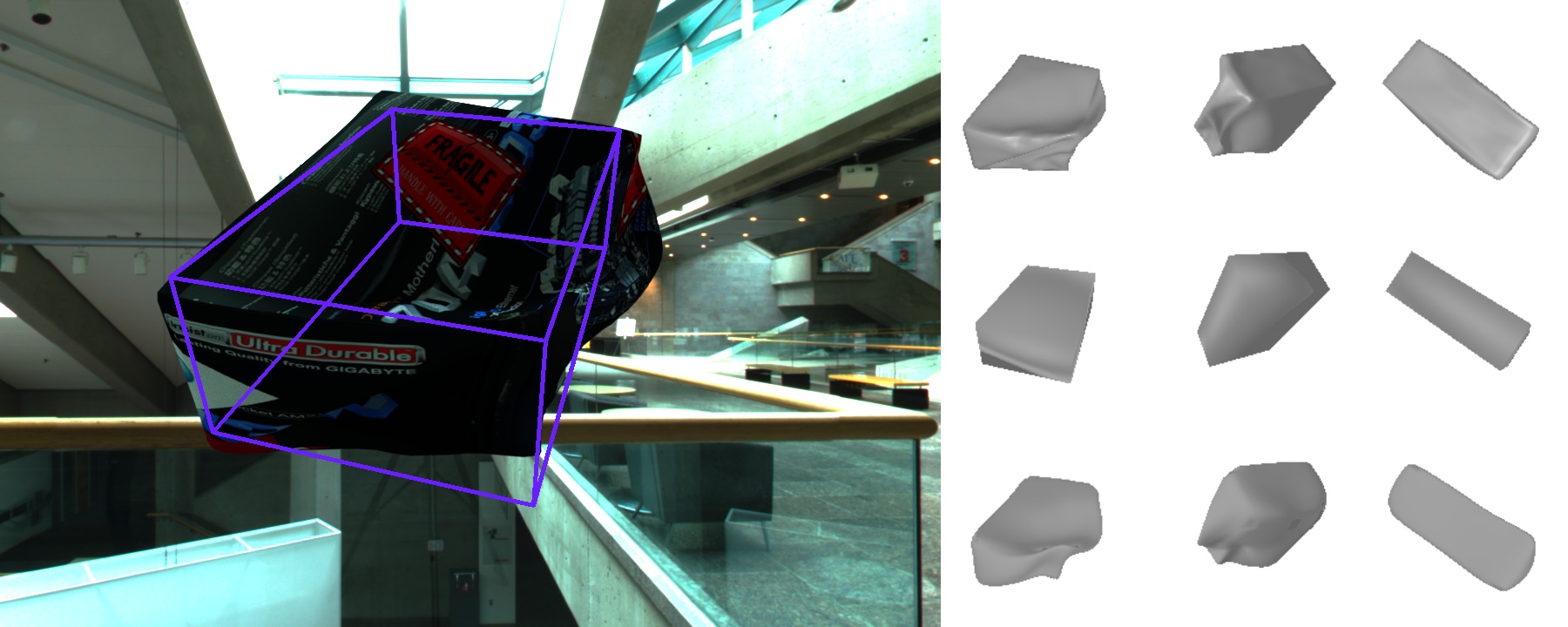}
        \includegraphics[width=\linewidth]{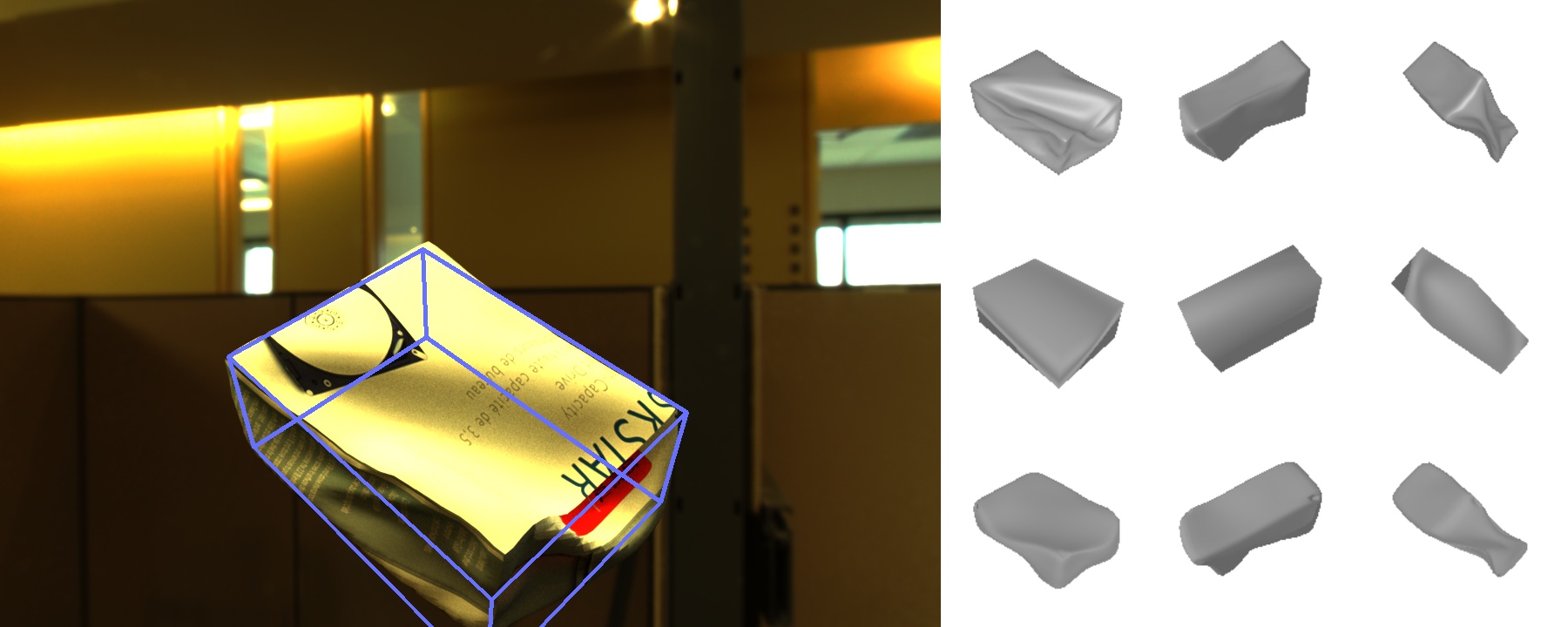}
        \caption{\cuberefine{} (ours)}
    \end{subfigure}
    \begin{subfigure}[b]{0.47\linewidth}
        \centering
        \includegraphics[width=\linewidth]{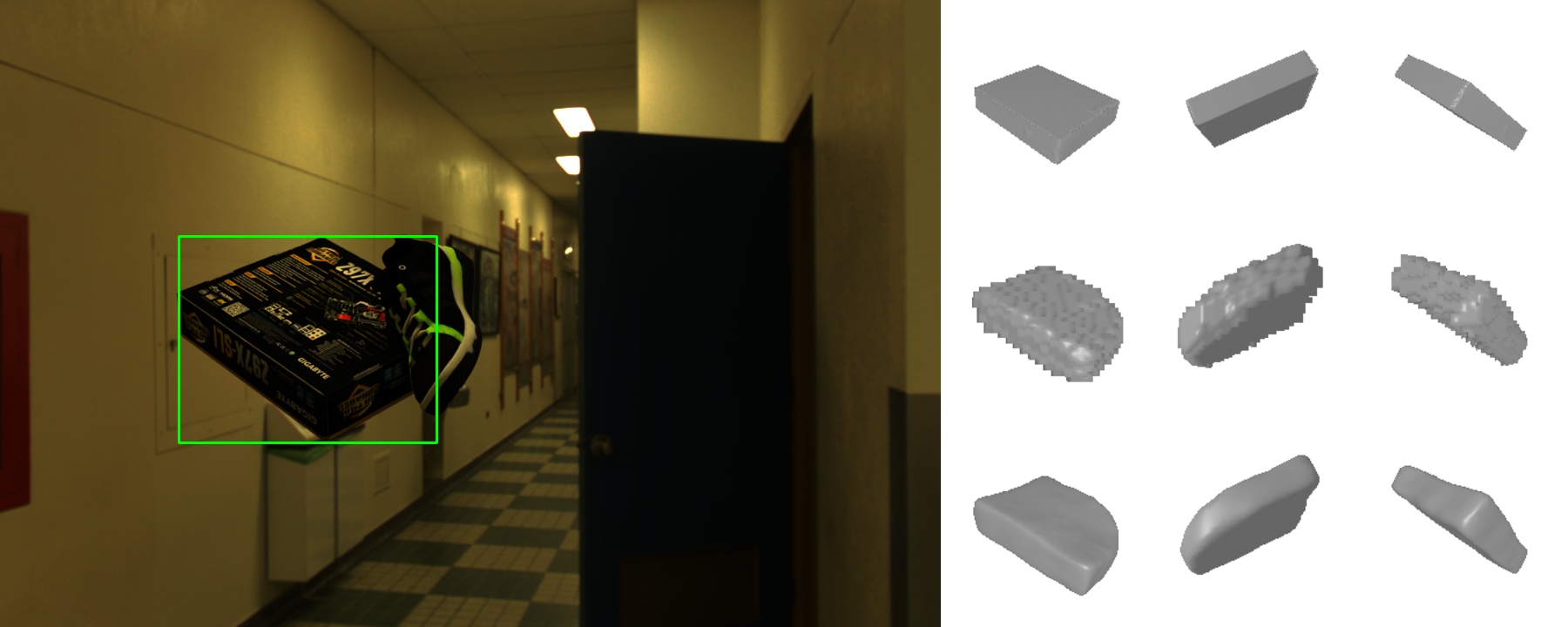}
        \includegraphics[width=\linewidth]{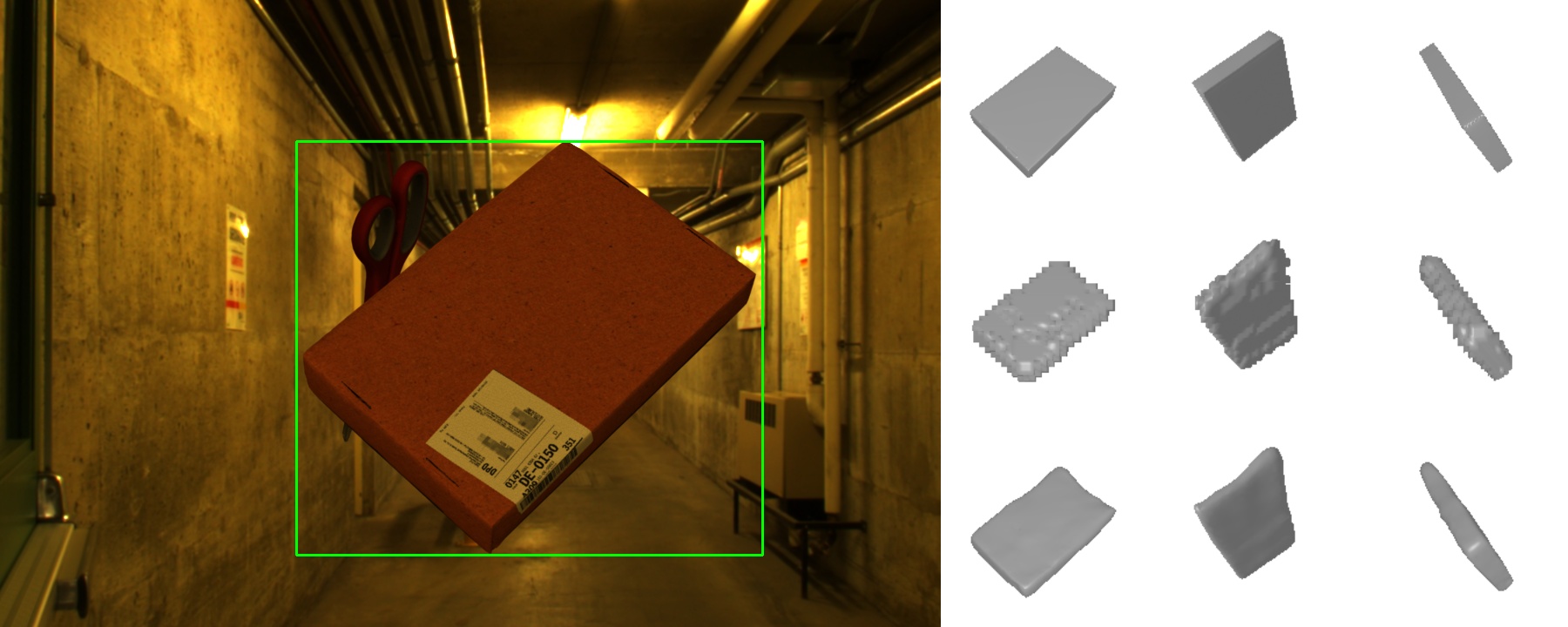}
        \includegraphics[width=\linewidth]{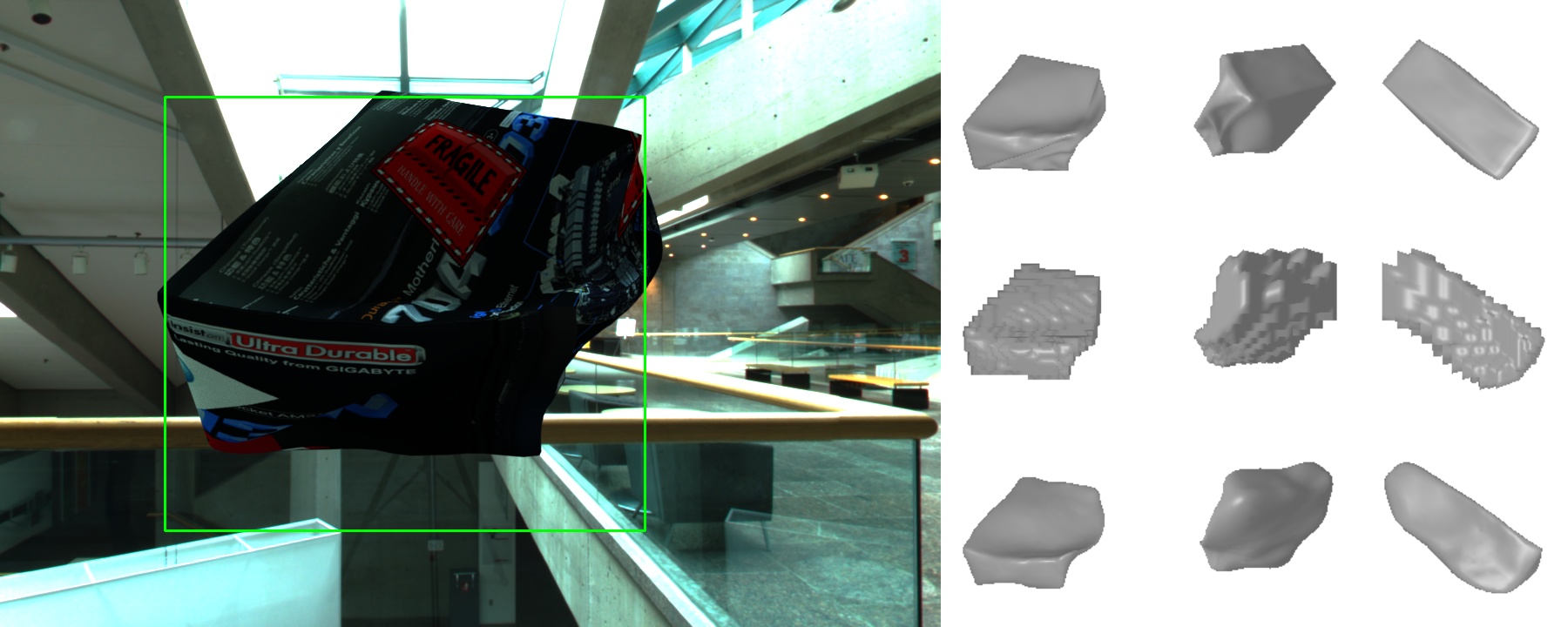}
        \includegraphics[width=\linewidth]{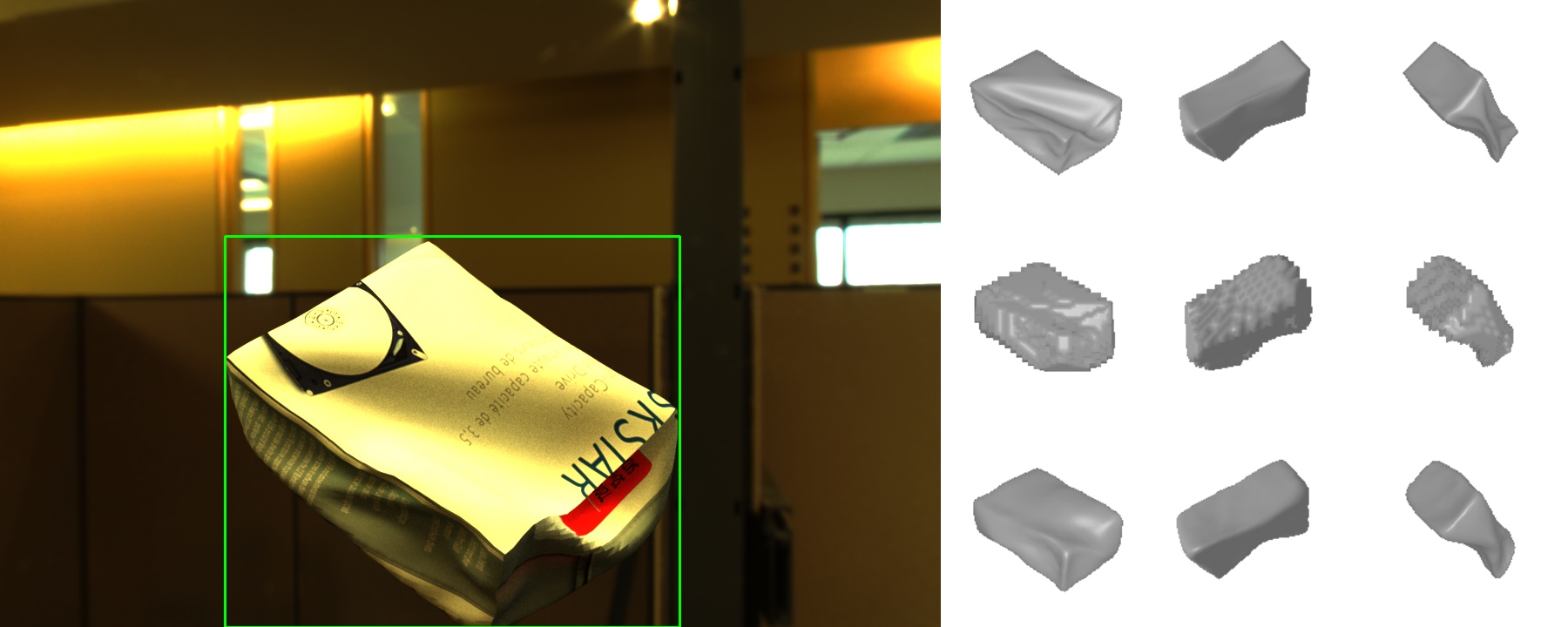}
        \caption{\meshrcnn{} \cite{gkioxariMeshRCNN2019}}
    \end{subfigure}
    \caption{
        Exemplary qualitative results for synthetic intact (row 1, 2) and damaged parcels (row 3, 4) for (a) \cuberefine{} and (b) \meshrcnn{}. 
        Per model, the input image with the detected 2D or 3D bounding box is shown on the left, and a $3\times3$ grid of mesh reconstructions on the right.
        Each column of the grid shows a different viewing angle, and the rows contain ground truth, 3D bounding box or voxelization (depending on the model) and refined mesh, respectively.
    }
    \label{fig:eval:synthetic}
\end{figure}

Considering 3D bounding box detection in the case of cuboid-shaped parcels, \cubercnn{} and \cuberefine{} perform best \wrt \meshap{75}, \cd{} and \nc{}, since they explicitly model cuboid-shaped objects.
Our additional mesh refinement increases performance compared to the base model \cubercnn{} by $2.8$ percentage points in \meshap{75}.
\meshrcnn{} still performs competitively, and the qualitative inspection (\cf \cref{fig:eval:synthetic}) suggests that differences mainly stem from difficulties in reconstructing the nonvisible, (self-)occluded parts of objects.
\cubercnn{} and \cuberefine{} do not suffer from this problem as much, since symmetry is imposed by the predicted 3D bounding box.

Considering only damaged parcels, we observe that predicting a voxel occupancy grid as done in \meshrcnn{} is advantageous.
\meshrcnn{} performs best in \meshap{} and \nc{}.
Despite high-quality 3D object detection, as suggested by the results in \cref{table:eval:cube}, \cuberefine{} has difficulties to adopt to the fine-grained meshes of damaged parcels.
This is observed in the considerably lower \meshap{}.
However, the better \cd{} suggests that general alignment with the ground truth is very high for \cuberefine{}.
This can also be observed in qualitative samples as visualized in \cref{fig:eval:synthetic} and might be caused by the symmetry the 3D bounding box imposes for (self-)occluded object parts.
\cubercnn{} performs poorly, as it only predicts 3D bounding boxes and thus, cannot take the damages into account.

\subsection{Real Data}
\label{sec:eval:real}

For the evaluation of the usability of our approach in \realworld{} applications, we use a dataset of parcels photos in various environments \cite{naumannScrapeCutPasteLearn2022}.
The dataset was generated using a custom camera rig to capture images with a depth and a stereo camera at the same time.
The depth information is then used to automatically generate annotations, which can be projected onto the stereo images.
The validation dataset comprises $96$ and the test dataset $297$ images.
Note, that it contains only normal parcels, since the annotation generation process was automated using the assumption of a cuboid shape.

Shape reconstruction on real images of cuboid-shaped parcels is more challenging due to the reality gap, as can be seen from the generally lower performance in \cref{table:eval:mesh}.
\cuberefine{} performs best despite having a low 2D bounding box detection precision compared to \meshrcnn{}.
Note, that \meshap{}, \cd{} and \nc{} were computed on meshes normalized within a unit cube, due to the scale ambiguity.
While \cubercnn{} is able to estimate scale, our synthetic training data is generated randomly, and thus, does not allow a scale transfer to the real world.

\begin{figure}
    \begin{subfigure}[b]{\linewidth}
        \centering
        \includegraphics[width=.47\linewidth]{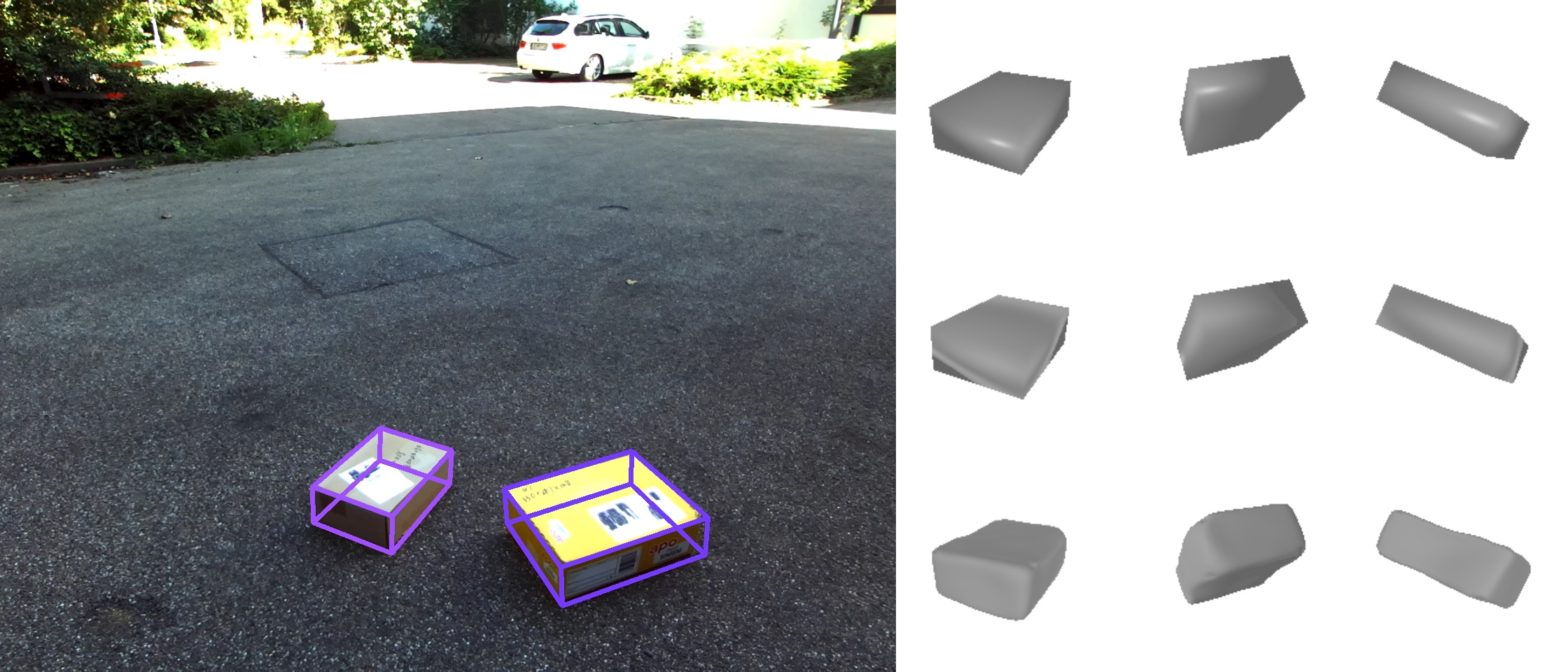}
        ~
        \includegraphics[width=.47\linewidth]{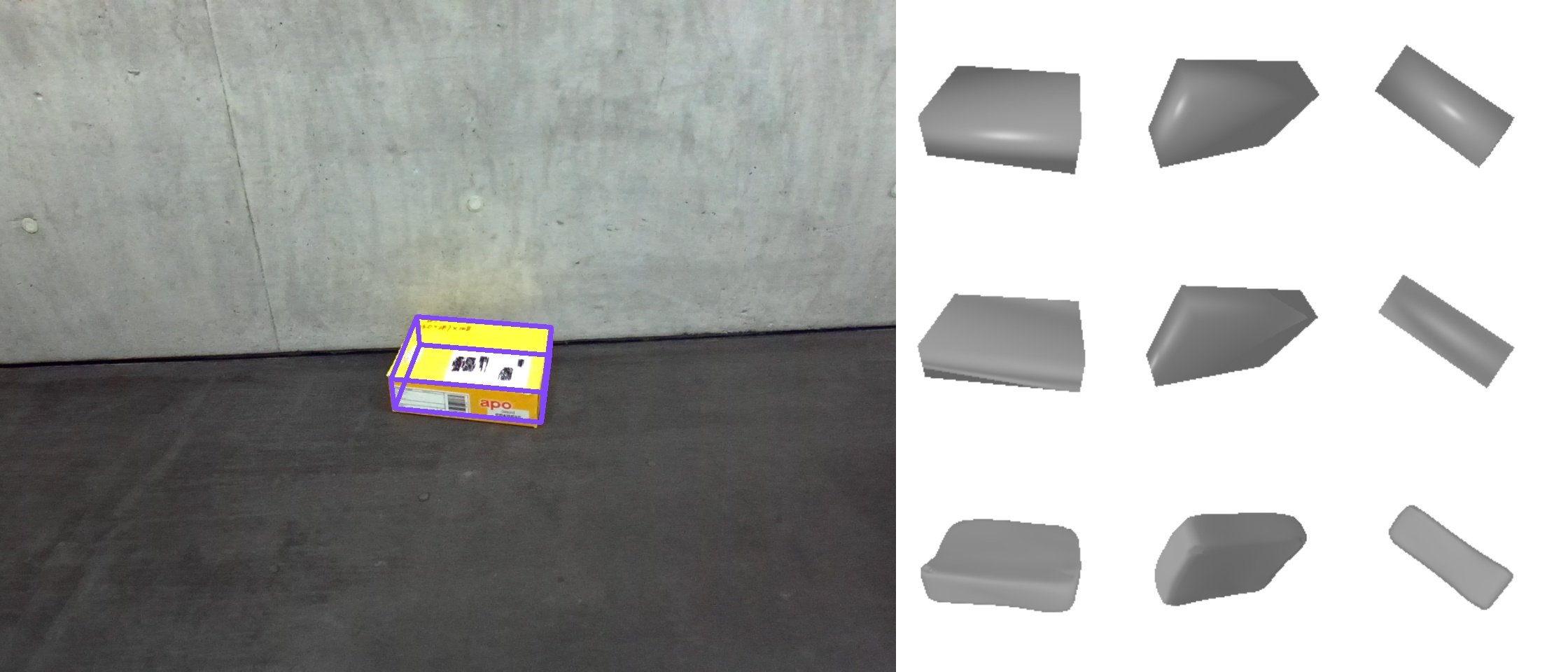}
        \caption{Successful Reconstructions}
        \label{fig:eval:real:a}
        \vspace{2ex}
    \end{subfigure}
    \begin{subfigure}[b]{\linewidth}
        \centering
        \includegraphics[width=.47\linewidth]{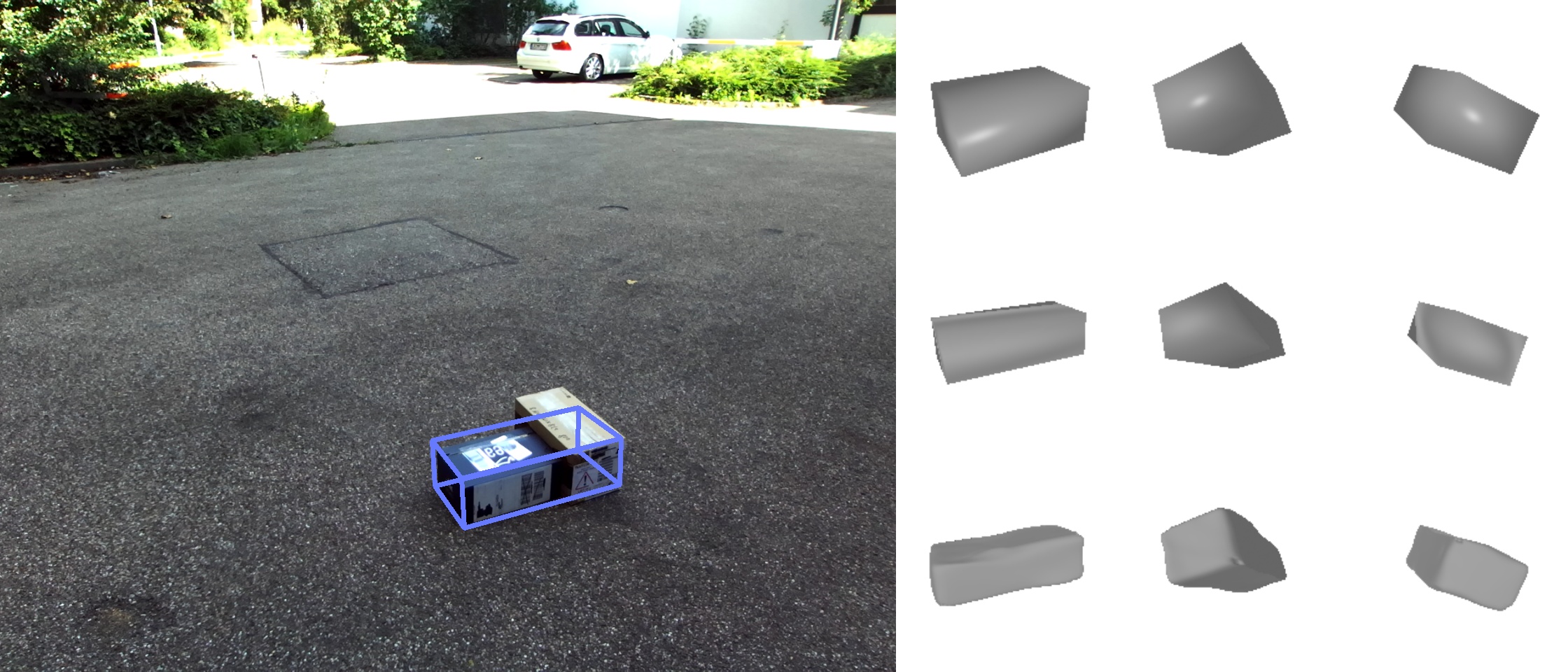}
        ~
        \includegraphics[width=.47\linewidth]{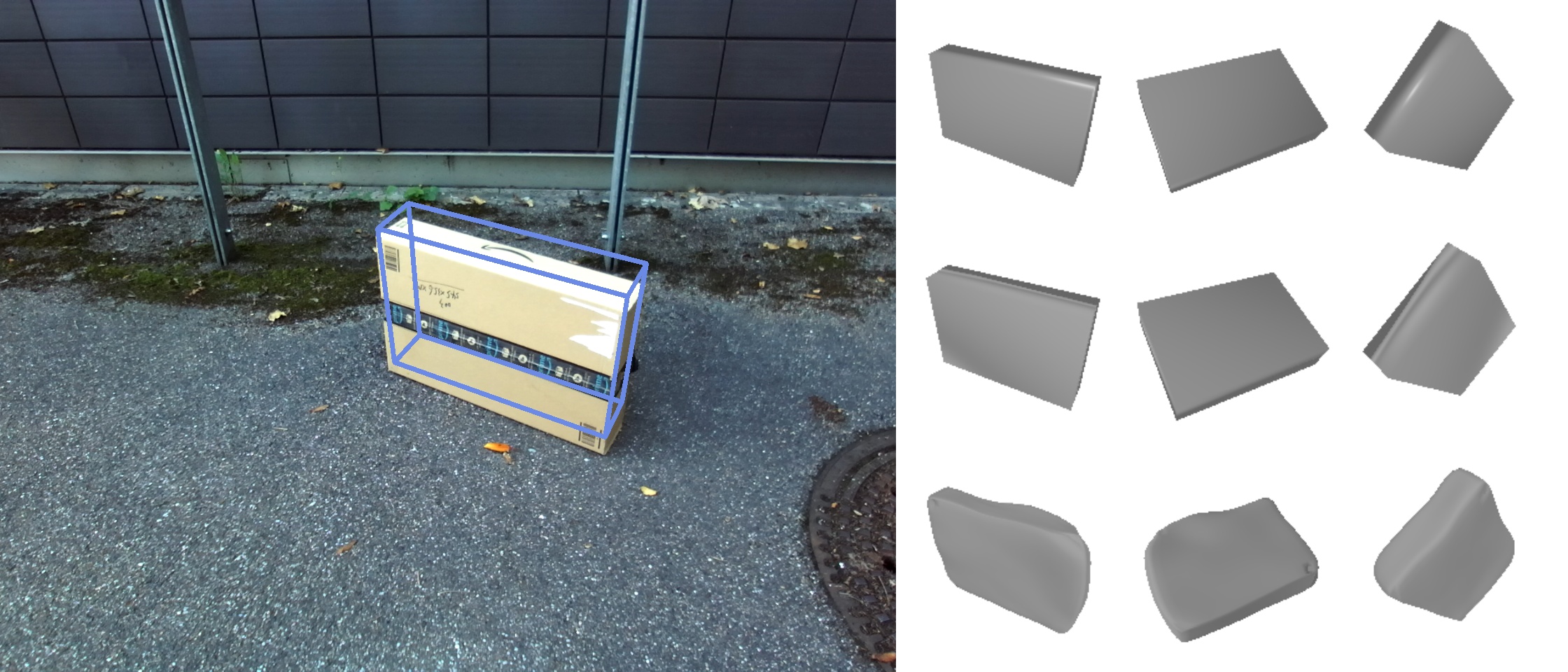}
        \caption{Problematic Cases}
        \label{fig:eval:real:b}
        \vspace{1ex}
    \end{subfigure}
    \begin{subfigure}[b]{\linewidth}
        \centering
        \includegraphics[width=.47\linewidth]{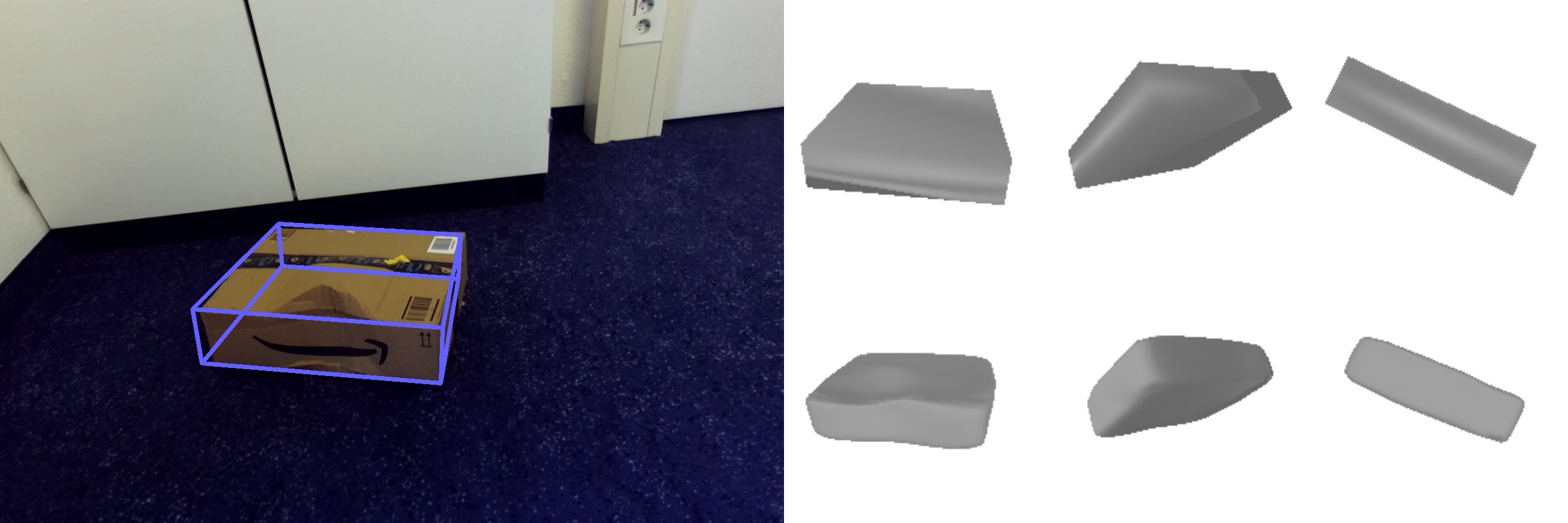}
        ~
        \includegraphics[width=.47\linewidth]{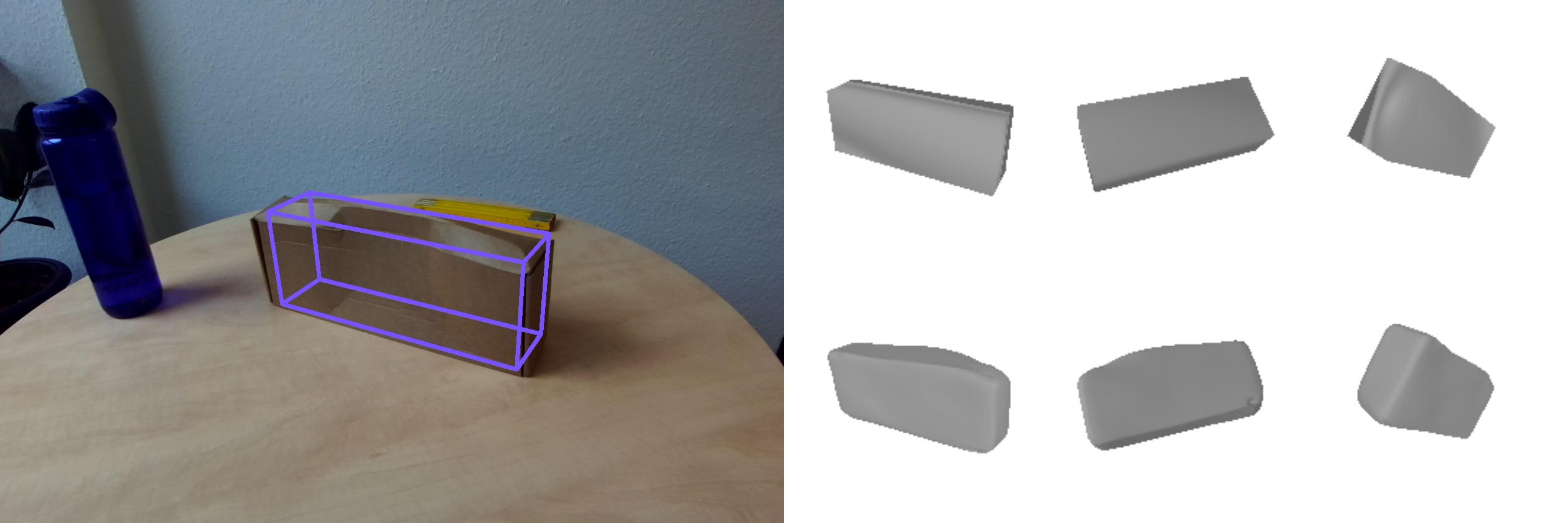}
        \caption{Damaged Parcels}
        \label{fig:eval:real:c}
        \vspace{1ex}
    \end{subfigure}
    \caption{
        Exemplary qualitative results for real parcels using \cuberefine{}. 
        We show the input image with the projected 3D bounding box on the left, and a $3\times3$ grid of mesh reconstructions on the right.
        Each column shows a different viewing angle, and the rows contain ground truth, 3D bounding box and refined mesh, respectively.
        Note, that for damaged parcels no ground truth is available.
    }
    \label{fig:eval:real}
\end{figure}

We present qualitative samples in \cref{fig:eval:real} and observe accurate reconstructions, when the object is localized correctly.
However, common error cases include not being able to distinguish nearby positioned parcels and inaccurate or missing localizations (\cf \cref{fig:eval:real:b}).
Since there are no \realworld{} datasets with full 3D annotations, we focus on brief insights into our qualitative inspection of damaged parcels.
The simulated deformation process that was presented in \cref{sec:data:blender} does not seem to represent the great variance of real-world deformations closely enough.
Thus, performance on real-world data is still limited as can be seen in \cref{fig:eval:real:c}.

\subsection{Applicability Summary}
\label{sec:eval:summary}

We summarize the advantages and limitations of our approach, and present brief insights into using damage quantification and tampering detection in practice.

\paragraph*{Advantages.}
We argue, that while \meshrcnn{} performs best in the case of damaged parcels, our approach is still advantageous for real-world application due to the following reasons:
(1) our approach is more lightweight and predicts both the current, potentially deformed shape of an object and its original shape at the same time.
This allows a direct 3D mesh comparison between the original and the deformed shape for damage quantification.
(2) The lower \meshap{} and better \cd{} compared to \meshrcnn{} suggest that our model represents the overall damage pattern well, however, is not as detailed as \meshrcnn{}.
We argue, that this is sufficient for damage pattern recognition in 3D, which is only enabled by our model.
(3) 3D bounding box detection enables using viewpoint invariant parcel side surface representations for tampering detection, as will be explained in the respective paragraph.

\paragraph*{Limitations.}
While \cuberefine{} has important advantages for \realworld{} use-cases, enabling reliable deployment in real-world scenarios is still challenging, presumably due to the constrained variance of deformations within \dsparcel{} and the domain shift caused by our training on synthetic data.
Furthermore, it is important to note that we focus on deformations of the packaging and do not treat other types of damages which frequently occur in practice (\eg water damage).
It is also not possible to reliably infer the impact of packaging deformations on the state of the transported good.
This information is essential to estimate economic damages.

\paragraph*{Damage Quantification.}
To utilize our model for automated deformation quantification and pattern recognition, metrics for 3D mesh comparison are necessary.
The change in volume between the original and current shape constitutes a simple metric that can be readily computed and interpreted.
However, mere volume analysis does not take the deformation location into account.
To remedy this, extending the axis-aligned pointcloud representation of the original 3D model by the per-point distance to the nearest neighbor of its potentially deformed version, and clustering in this 4D space can help to identify areas that underwent the strongest deformations.
Further clustering across parcel instances can provide insights into damage patterns.
Moreover, normalized voxel grid occupancy differences can be analyzed by considering the union of the voxelized meshes and subtracting their intersection.

\paragraph*{Tampering Detection.}
From the 3D bounding box output of \cuberefine{} we can infer the visible parcel side surfaces and project them back onto the image.
For each such parcel side surface, a perspective transformation can be applied to obtain normalized fronto-parallel views.
These representations have already been successfully used for tampering detection \cite{nocetiMulticameraSystemDamage2018} and re-identification \cite{ruiGeometryConstrainedCarRecognition2020}.
For tampering detection, recent advances in change detection \cite{shiChangeDetectionBased2020} could be leveraged.

\section{Conclusion}
\label{sec:conclusion}

In this work, we present an approach for simultaneous detection and shape reconstruction of intact and damaged parcels from single RGB images, called \cuberefine{}.
We extend \cubercnn{} \cite{brazilOmni3DLargeBenchmark2023} by an iterative mesh refinement to benefit from a cuboid prior, while at the same time enabling adjustments for damages in parcels.
To overcome the lack of existing datasets, we also introduce \dsparcel{}, a novel synthetic dataset of intact and damaged parcel images with full 3D annotations that is suitable for applications in transportation logistics and warehousing.
To generate the dataset, we leverage selected data from the \glsfirst*{gso} dataset \cite{downsGoogleScannedObjects2022}.
We combine these with models of damaged parcels that were generated using physics-based simulations and a new dataset of synthetic cardboard textures.

Our approach outperforms existing baselines for intact parcels and performs competitively for damaged parcels.
While \meshrcnn{} \cite{gkioxariMeshRCNN2019} yields the best results for the case of damaged parcels, our approach is the only one directly enabling deformation assessment %
and tampering detection. %
The results of our approach are promising to help identifying systematic mishandling of parcels, especially in scenarios where only simple sensor data is available, as during last mile delivery to a client for example.
However, the reliable deployment in \realworld{} scenarios is still challenging.
More diverse and realistic shape deformation types within the dataset and \realworld{} training data are promising improvements.

~\\
{
\textbf{Acknowledgement: }
We thank Zeyu Wang for helping with the implementation of the rendering pipeline and for fruitful discussions in the early stages of the project.
}

\printbibliography
\end{document}